\documentclass[conference]{IEEEtran}
\IEEEoverridecommandlockouts
\usepackage{cite}
\usepackage{amsmath,amssymb,amsfonts}
\usepackage{algorithmic}
\usepackage{graphicx}
\usepackage{textcomp}
\usepackage{xcolor}
\usepackage{comment}
\usepackage{tikz}
\usepackage{subcaption}
\usepackage{multirow}
\usepackage{comment}
\usepackage{booktabs}
\usetikzlibrary{positioning, arrows.meta, calc}

\def\BibTeX{{\rm B\kern-.05em{\sc i\kern-.025em b}\kern-.08em
    T\kern-.1667em\lower.7ex\hbox{E}\kern-.125emX}}

\definecolor{ageBlue}{rgb}{0.1, 0.1, 0.9}
\definecolor{bpCyan}{rgb}{0.1, 0.6, 0.7}
\definecolor{ldlRed}{rgb}{0.8, 0.2, 0.1}
\definecolor{hdlTeal}{rgb}{0.1, 0.5, 0.5}
\definecolor{diabPurple}{rgb}{0.5, 0.1, 0.8}
\definecolor{hdBlue}{rgb}{0.1, 0.4, 0.8}
\definecolor{fhRed}{rgb}{0.9, 0.1, 0.1}
\definecolor{smkOrange}{rgb}{1.0, 0.5, 0.0}
\definecolor{bmiMagenta}{rgb}{0.9, 0.1, 0.7}
\definecolor{exGreen}{rgb}{0.3, 0.6, 0.1}
\definecolor{sexCyan}{rgb}{0.2, 0.8, 0.9}
\definecolor{shHist}{rgb}{0.1, 0.1, 0.1}

\definecolor{ageBlue}{rgb}{0.1, 0.1, 0.9}
\definecolor{bpCyan}{rgb}{0.1, 0.6, 0.7}
\definecolor{ldlRed}{rgb}{0.8, 0.2, 0.1}
\definecolor{hdlTeal}{rgb}{0.1, 0.5, 0.5}
\definecolor{diabPurple}{rgb}{0.5, 0.1, 0.8}
\definecolor{hdBlue}{rgb}{0.1, 0.4, 0.8}
\definecolor{fhRed}{rgb}{0.9, 0.1, 0.1}
\definecolor{smkOrange}{rgb}{1.0, 0.5, 0.0}
\definecolor{bmiMagenta}{rgb}{0.9, 0.1, 0.7}
\definecolor{exGreen}{rgb}{0.3, 0.6, 0.1}
\definecolor{sexCyan}{rgb}{0.1, 0.85, 0.9}
\definecolor{histBrown}{rgb}{0.5, 0.21, 0.05}

\begin{document}

\title{Combining Relevant Expert Knowledge Using Mixtures of Feedback Fuzzy Cognitive Maps\\
%{\footnotesize \textsuperscript{*}Note: Sub-titles are not captured for https://ieeexplore.ieee.org  and should not be used}
}

\author{ \IEEEauthorblockN{Olaoluwa Adigun}
 \IEEEauthorblockA{ \textit{School of Computing \& Information Sciences}\\ \textit{Florida International University} \\
\textit{ Miami, USA}\\ 
olaadigu@fiu.edu }
 \and 
 \IEEEauthorblockN{Akash Kumar Panda} 
 \IEEEauthorblockA{ \textit{Department of Electrical and Computer Engineering}\\ 
 \textit{University of Southern California} \\
\textit{Los Angeles, USA}\\ akashpan@usc.edu } 
\and 
% \vspace{0.1em}
  \IEEEauthorblockN{\hspace{5cm}Bart Kosko} 
\IEEEauthorblockA{ \hspace{5cm}\textit{Department of Electrical and Computer Engineering}\\ 
%\hspace{5cm} \textit{and}\\
\hspace{5cm} \textit{Gould School of Law}\\
\hspace{5cm} \textit{University of Southern California}\\
 \hspace{5cm}\textit{Los Angeles, USA}\\
  \hspace{5cm}kosko@usc.edu } }

\begin{comment}
\author{\IEEEauthorblockN{Olaoluwa Adigun}
\IEEEauthorblockA{\textit{ School of Computing \& Information Sciences } \\
\textit{Florida International University}\\
Miami, USA \\
olaadigu@fiu.edu}
\and
\IEEEauthorblockN{Akash Kumar Panda}
\IEEEauthorblockA{\textit{Department of Electrical and Computer Engineering} \\
\textit{University of Southern California}\\
Los Angeles, USA \\
akashpan@usc.edu}
\and
\IEEEauthorblockN{Bart Kosko}
\hfill \IEEEauthorblockA{\textit{Department of Electrical and Computer Engineering} \\
\textit{Gould School of Law}\\
\textit{University of Southern California}\\
Los Angeles, USA \\
kosko@usc.edu}
%\and
%\IEEEauthorblockN{\hspace{2.0in} Bart Kosko}
%\IEEEauthorblockA{\hspace{2.1in} \textit{Department of Electrical and Computer Engineering} \\
%\textit{\hspace{1.9in} University of Southern California}\\
%\hspace{2.0in}Los Angeles, USA \\
%\hspace{2.0in} kosko@usc.edu}
}
\end{comment}

%\vspace{-0.3in}

\maketitle

\begin{abstract}
A neuro-symbolic system should combine only relevant answers from expert agents or other knowledge sources.
Using only relevant experts for a given input query becomes ever more important for accuracy and scalability as the knowledge network grows and connects ever more knowledge sources.
We extend the standard definition of relevance from the legal theory of evidence: An answer is relevant if and only if it is both probative and material for the matter at hand given the knowledge available.
We apply this relevancy scheme to mixing the feedback fuzzy cognitive maps (FCMs) of real or artificial experts.
The expert FCM directed causal graphs mix or combine into a common knowledge network.
We compare the equilibrium limit-cycle predictions or answers of the experts to the answers of the mixture FCM and then estimate the probative status of answers by their difference or variation.  
A new Bayes-like theorem de-mixes the knowledge mixture into corresponding posterior-like FCMs that also allow relevancy-based comparisons.
The result is a final set of relevancy mixing weights that produces a relevancy-mixed FCM knowledge network for the given query.
So different queries produce different relevancy-mixed FCMs.
We demonstrate this relevancy approach using generated synthetic FCMs from an actual 13-node expert FCM that models the causes and feedbacks of ischemic-stroke risk.
\end{abstract}


\section{Relevant Answers = Probative and Material}

\begin{figure*}[htbp]
\centerline{\includegraphics[width=0.9\textwidth]{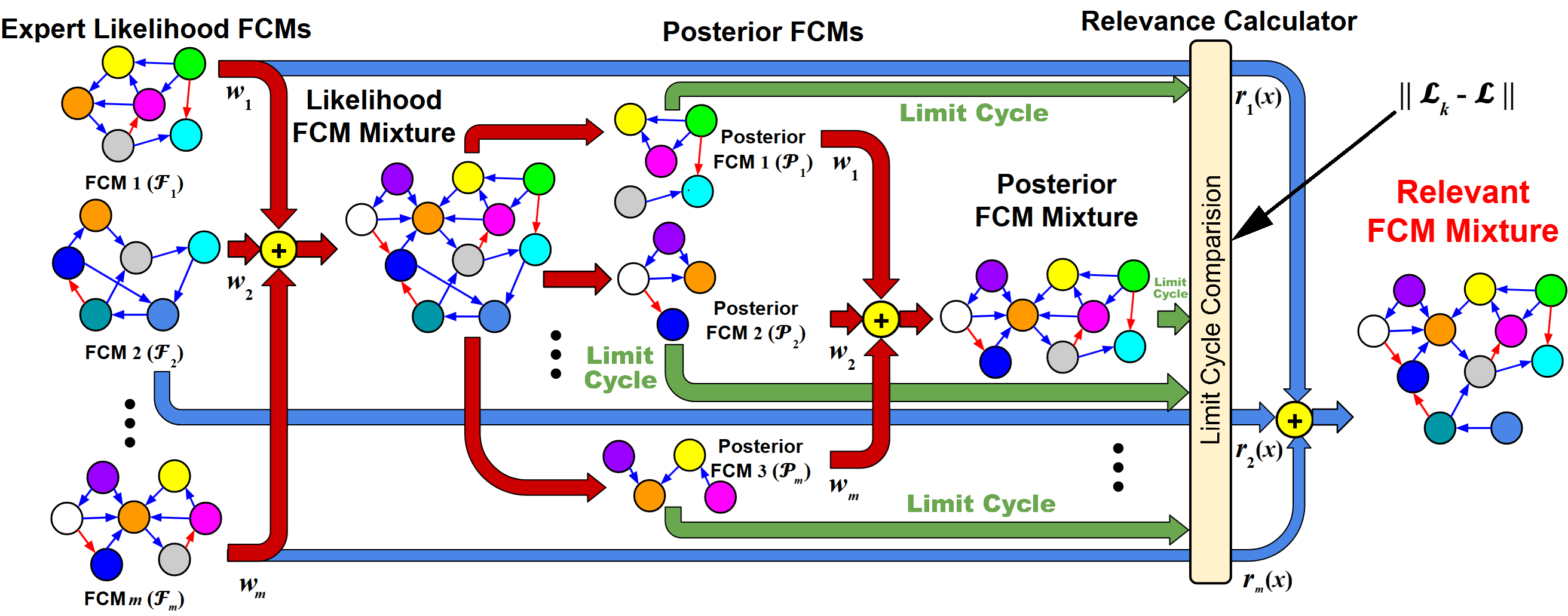}}
\caption{\small{Fuzzy Cognitive Map (FCM) mixture of relevant experts. 
The expert likelihood FCMs $\mathcal{F}_k$ on the left mix through convex combination with prior mixing weights $w_k$ to give the likelihood FCM mixture $\mathcal{F}$. 
The likelihood FCM mixture gives Bayesian posterior-like FCMs $\mathcal{P}_k$ that produce limit cycle equilibria $\mathcal{L}_k$. 
These posterior FCMs also mix with the same prior mixing weights $w_k$ to give the posterior FCM mixture $\mathcal{P}$ with its limit cycle $\mathcal{L}$. 
The convex relevancy weights $r_k(x)$ for query $x$ measure the relevance of the likelihood FCMs by comparing the posterior FCMs' limit cycles $\mathcal{L}_k$ to the limit cycle $\mathcal{L}$ of the posterior FCM mixture through the norm of their difference.
The likelihood FCMs mix with the final relevance weights $r_k(x)$ to produce the final relevant FCM given input $x$. }}
\label{fig:Relevancy}
%\vspace{-0.18in}
\end{figure*}

We address a basic epistemic question for combined neuro-symbolic agentic systems:  When is an expert's answer \emph{relevant} to a question?
A knowledge network should combine only relevant answers or evidence to a query and weight them as accurately as the facts permit.

Figure~\ref{fig:Relevancy} shows the proposed architecture of such a relevance-based combination of agentic experts modeled as feedback fuzzy cognitive maps (FCMs).
FCMs are signed cyclic knowledge graphs with weighted or ``fuzzy" directed causal edges \cite{glykas2010fuzzy, kosko1986fuzzy}.
Given expert FCMs serve as likelihood-like operators in a Bayesian scheme for mixing and de-mixing the growing causal knowledge base.
The first step mixes the $m$ likelihood FCMs with prior weights $w_1, \ldots , w_m$ into the combined likelihood FCM.
This mixing step accounts for missing causal nodes in the FCMs by inserting zero-padded rows and columns for the missing nodes.
This trick gives $m$-many $n$-by-$n$ matrices of directed causal edges that conform for addition.
A new Bayes-like theorem produces $m$ corresponding posterior-like FCMs that also mix with the prior weights. 
We then compare in norm the equilibrium predictions (dynamical limit cycles and fixed points) of the likelihood and posterior FCMs.
This leads to a set of $m$ normalized \emph{relevancy} mixing weights $r_1, \ldots, r_m$ for the final relevancy-based FCM mixture.
We answer the relevance question itself by using the standard definition of relevance from the law of evidence \cite{strong1999mccormick}:  An expert's answer is \emph{relevant} to the case or problem at hand if and only if the answer is both  \emph{probative} and \emph{material}.

A common way of stating this rule for proposed evidence in a dispute is that the evidence is relevant iff it tends to prove or disprove a fact of consequence to the dispute or litigation.
\emph{Federal Rule of Evidence} 401 expressly breaks the same definition into its conjoined probative and material parts:  ``Evidence is relevant if:  (a) it has any tendency to make a fact more or less probable than it would be without the evidence; and (b) the fact is of consequence in determining the action."
Section 210 of the \emph{California Evidence Code} likewise defines relevant evidence as ``having any tendency in reason to prove or disprove any disputed fact that is of consequence to the determination of the action." 

This two-part answer gives a simple multiplicative definition of the relevancy $r(x)$ of an answer or evidence $x$ in terms of its materiality $\tilde{m}(x)$ and its probativeness $p(x)$ if we interpret the conjunctive \emph{and} as a product:
\begin{equation}
	r(x) =  p(x) \ \tilde{m}(x)    \ .   \label{relevance}
\end{equation}
The system functionals $p$ and $\tilde{m}$ map the same pattern vector space $X$ to the nonnegative real numbers.
This law-inspired definition of relevance differs in kind from earlier definitions in information retrieval based on topicality or the ``aboutness" of documents \cite{schamber1990}.
It also reflects the black-letter law of evidence that only relevant evidence is admissible because irrelevant expert answers produce zero relevancy weights that exclude the expert from the mixture for that specific question.

Materiality with experts can be as simple as common subject matter or domain focus.
The law finds evidence material if it relates to a disputed element of a cause of action.
McCormick calls this the ``fit between the evidence and the case" \cite{strong1999mccormick}.
That fit can include whether the defendant in a negligence action could foresee the harm to the plaintiff or whether the defendant in a criminal case had the specific intent to do what the state alleges that he did.

This paper explores mixed knowledge networks from generated synthetic medical experts who focus on ischemic strokes based on an actual ischemic-stroke FCM \cite{khodadadi2019fuzzy}.
So we impliedly set the materiality measure $\tilde{m}$ to unity for all stroke experts involved:  $\tilde{m}(x) = 1$ for all pattern state vectors $x$.
We thus likewise dismiss all other non-stroke experts with $\tilde{m}(x) = 0$ for all $x$ and leave a more detailed exploration of expert materiality to future work.

Evidence is probative of a question if it tends to change its proof status.
McCormick defines probativeness as the ``tendency of evidence to establish the proposition that it is offered to prove" \cite{strong1999mccormick}.
A simple if idealized case is when evidence $E$ changes the unconditional probability $P(H)$ of a hypothesis $H$:  $E$ is probative for $H$ iff either $P(H) < P(H|E)$ or $P(H) > P(H|E)$.
So the magnitude $|P(H) - P(H|E)|$ measures the probativeness of $E$.
This corresponds to the Bayesian ``surprise" when going from the prior $P(H)$ to the posterior $P(H|E)$ and measures thereof \cite{itti2005bayesian}.
We can also use likelihood ratios to estimate probativeness or other metrics based on probability or belief.

Below we estimate the probative content of an expert answer by how that answer differs or varies from the answer of the mixture FCM that represents a type of working consensus of the experts.  
This involves comparing metrical difference between each expert's equilibrium-attractor answer with that of the likelihood or posterior mixture FCM.  
Normalizing gives the final relevancy mixing weights $r_1(x), \ldots, r_m(x)$ for the given query or question $x$ and where again we assume here that $m(x) = 1$ for all the subject-matter experts.
We also assume for simplicity that all subject-matter experts have the same unit credibility weight $\tilde{c}(x) = 1$.
A spectrum of credibility weights can in practice dampen or enhance how the expert's variability contribute to the relevancy computations.

The next sections review how FCM dynamical systems infer and mix along with the new Bayes-like posterior structures for de-mixing combined FCMs. 
Then we randomly generate 7 synthetic experts from the \emph{de facto} ground-truth FCM on ischemic stroke and then produce the final relevancy-mixed stroke FCM and its equilibrium predictions.
The main computations involve only vector-matrix and pseudo-inverse computations.
The causal-edge matrices in practice are often sparse matrices because of the zero-padding technique that augments them for conformable mixing.

\section{Overview of Fuzzy Cognitive Maps}
\label{sec:fcms}

FCMs model causal dynamical systems as directed weighted graphs \cite{kosko1986fuzzy, osoba2017fuzzy, ziv2018potential, glykas2010fuzzy, papageorgiou2013fuzzy, stach2010divide ,taber2007quantization, panda2025causal, panda2025agentic}. 
The FCM concept nodes describe the causal variables in the dynamical system and the directed edges describe the causal relationships between those nodes. 
FCMs allow feedback and therefore converge to nontrivial equilibria like limit cycles. 
FCMs model dynamical systems by approximating their underlying maps from initial states to corresponding equilibria.

\subsection{Causal Edge Matrix}

The directed edges of the FCM describe the causal relationships between concept nodes. 
An edge from the $i^\text{th}$ concept node $C_i$ to the $j^\text{th}$ concept node $C_j$ means ``$C_i$ causes $C_j$''. 
The fuzzy edge-weights describe partial causality. 
The edge weight $e_{ij} \in [-1,1]$ on the edge from the $i^\text{th}$ node to the $j^\text{th}$ node gives the degree to which $C_i$ causes $C_j$: $e_{ij} = \texttt{Degree}(C_i \rightarrow C_j)$.
%\begin{align}
 %   e_{ij} = Degree(C_i \rightarrow C_j).
%\end{align}
A positive $e_{ij}$ shows that $C_j$ increases when $C_i$ increases and a negative $e_{ij}$ means that $C_j$ decreases if $C_i$ increases. 
The weight $e_{ij}$ is zero when there is no causal edge between $C_i$ and $C_j$. 
The magnitude of $e_{ij}$ is high if there is a strong causal relationship between $C_i$ and $C_j$. 
Low magnitude of $e_{ij}$ describes a weak causal relationship between $C_i$ and $C_j$.

\begin{figure}[htbp]
    \centering
    \begin{subfigure}[t]{0.48\textwidth}
   % \vspace{0.20in}
    \centering
 	\includegraphics[width=\linewidth]{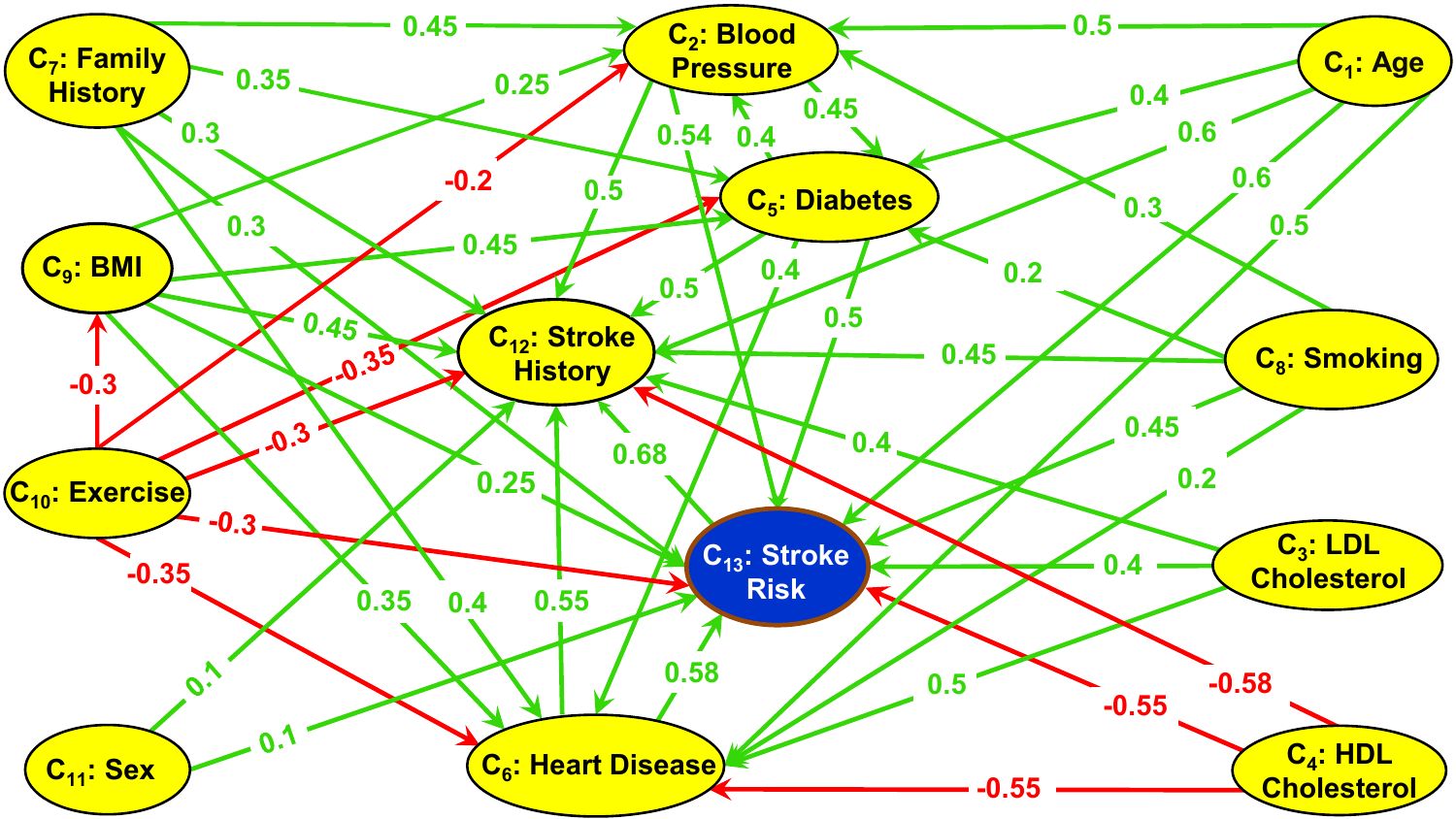}
    \caption{ Fuzzy Cognitive Map (FCM)}
    \label{fig:ischemic_stroke_fcm_web}
    \end{subfigure}
    \begin{subfigure}[t]{0.4\textwidth}
    \vspace{0.20in}
    \centering
 	\includegraphics[width=0.95\linewidth]{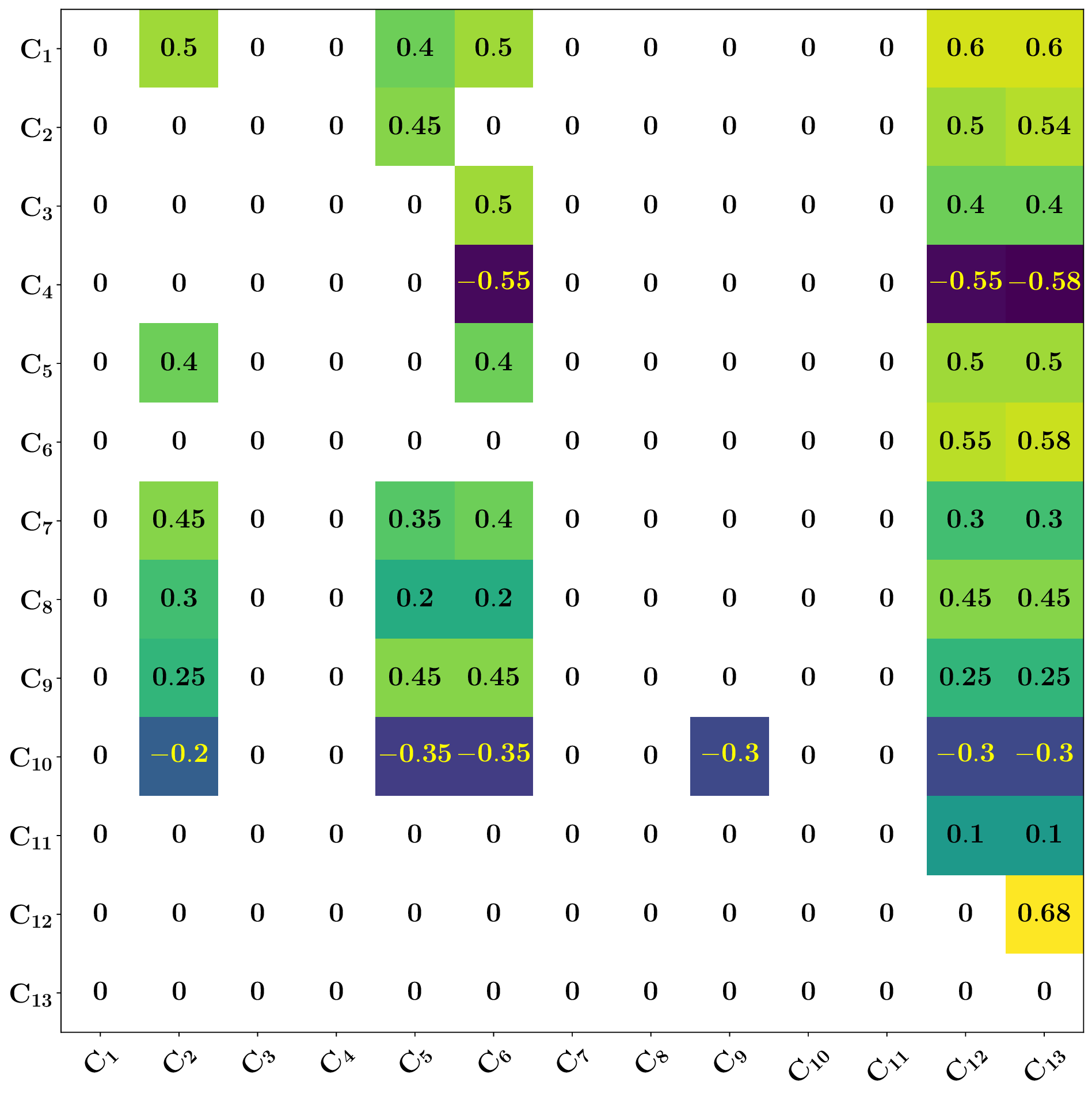}
 	\caption{ Causal edge matrix or fuzzy directed edge values}
    \label{fig:ischemic_stroke_edge_wgt_matrix}
    \end{subfigure}
    \hfill
     \begin{subfigure}[t]{0.48\textwidth}
    \vspace{0.00in}
    \centering
 	\includegraphics[width=0.9\linewidth,  trim=0 330 0 0, clip]{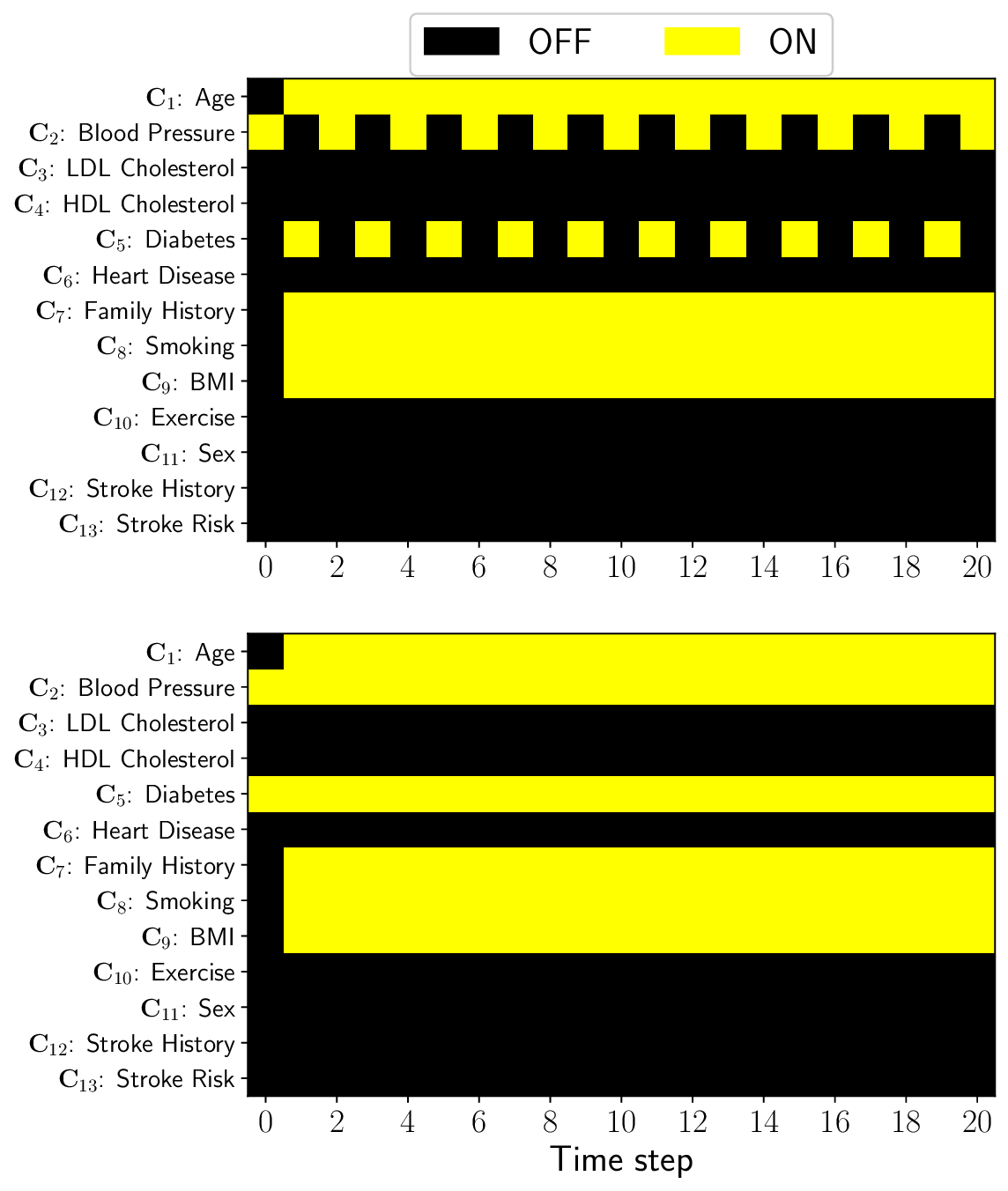}
 	\includegraphics[width=0.9\linewidth, trim=0 0 0 620, clip]{Figures/ischemic_stroke_att1.eps}
 	\caption{Equilibrium:  Limit cycle (top) and fixed point (bottom)}
    \label{fig:ischemic_stroke_equilibria}
    \end{subfigure}
    \caption{\small {Ischemic stroke FCM \cite{khodadadi2019fuzzy}: The 13-variable FCM represents the causal relationships and weights among the factors that affect the risk of ischemic stroke.
    (a) The FCM connects 13 demographic, clinical, and lifestyle causal variables using signed weighted edges. 
    Green edges denote positive causal influence and red edges denote negative influence on stroke risk ${\bf C}_{13}$.
    (b) Causal edge matrix $E$: directed partial causal edge $e_{ij}$ denotes the degree that ${\bf C}_{i}$ affects ${\bf C}_{j}$.
    (c) Plot of FCM equilibrium from an initial state.
The trajectory settles into a limit cycle where blood pressure ${\bf C}_2$ oscillates with diabetes ${\bf C}_5$.
%The bottom chart shows a trajectory that quicly converges to the fixed point with active nodes age, blood pressure, diabetes, family history, smoking, and body mass index (BMI).
}
    %Fuzzy Cognitive Map representing causal relationships and weights among stroke-related risk factors. Edges are colored by their source concept; negative weights denote inhibitory influence. Fuzzy Cognitive Map representing causal relationships and weights among ischemic stroke-related risk factors.
    }
    \label{fig:ischemic_stroke_fcm}
   % \vspace{-0.28in}
\end{figure}
An $n$-by-$n$ matrix $E$ describes all the directed weighted edges of an $n$-node FCM. 
The edge weight $e_{ij}$ appears as the matrix element on the $i^\text{th}$ row and the $j^\text{th}$ column. 
The matrix element is zero if there is no edge between the corresponding node-pair.

\begin{figure*}[htbp]

 \begin{subfigure}[t]{0.24\textwidth}
     \vspace{0.10in}
    \centering
 	\includegraphics[width=\linewidth]{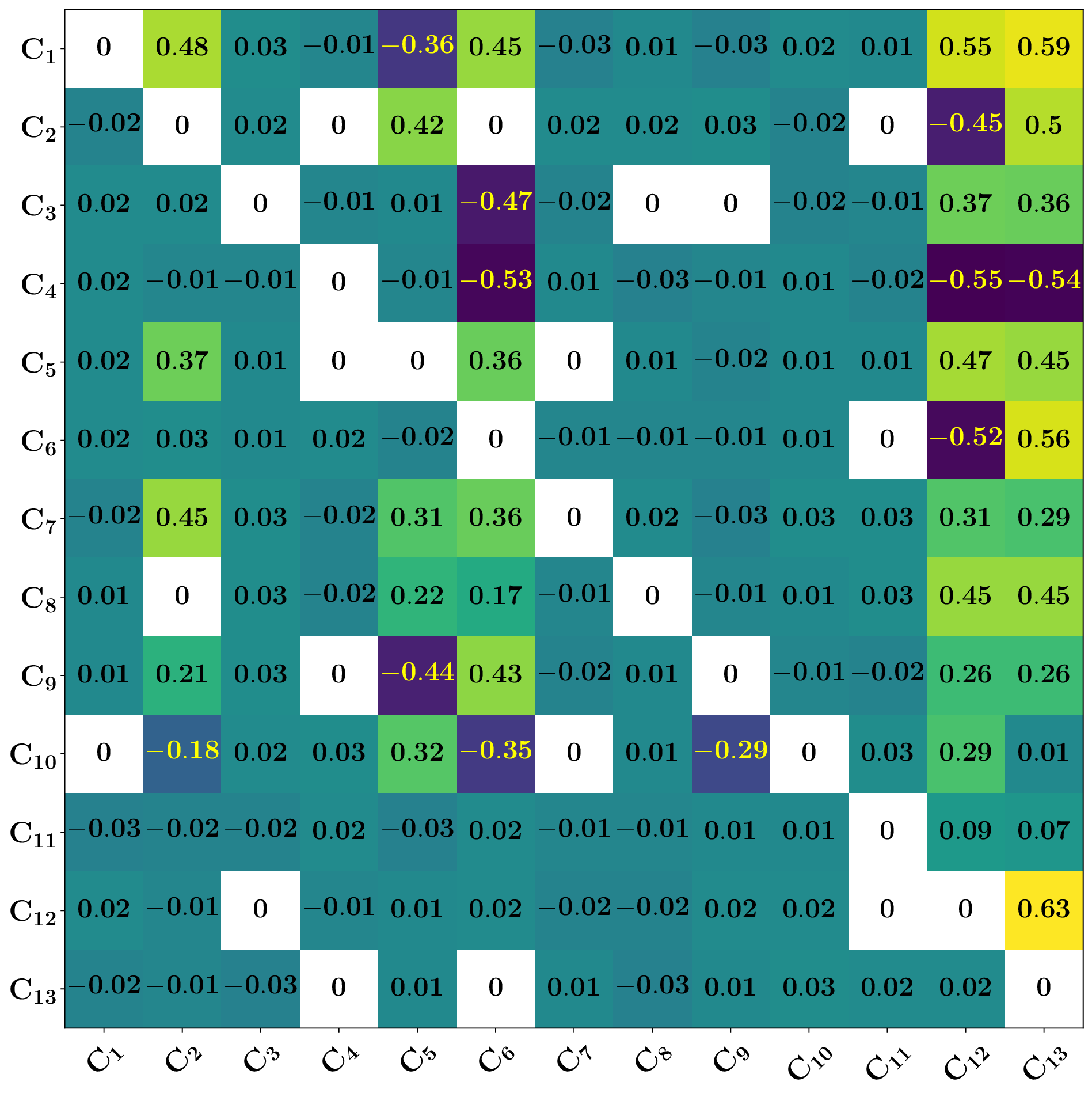}
 	\caption{Likelihood FCM $\mathcal{F}_1$}
    \label{fig:likelihood_1}
    \end{subfigure}  
    \hfill
 \begin{subfigure}[t]{0.24\textwidth}
     \vspace{0.10in}
    \centering
 	\includegraphics[width=\linewidth]{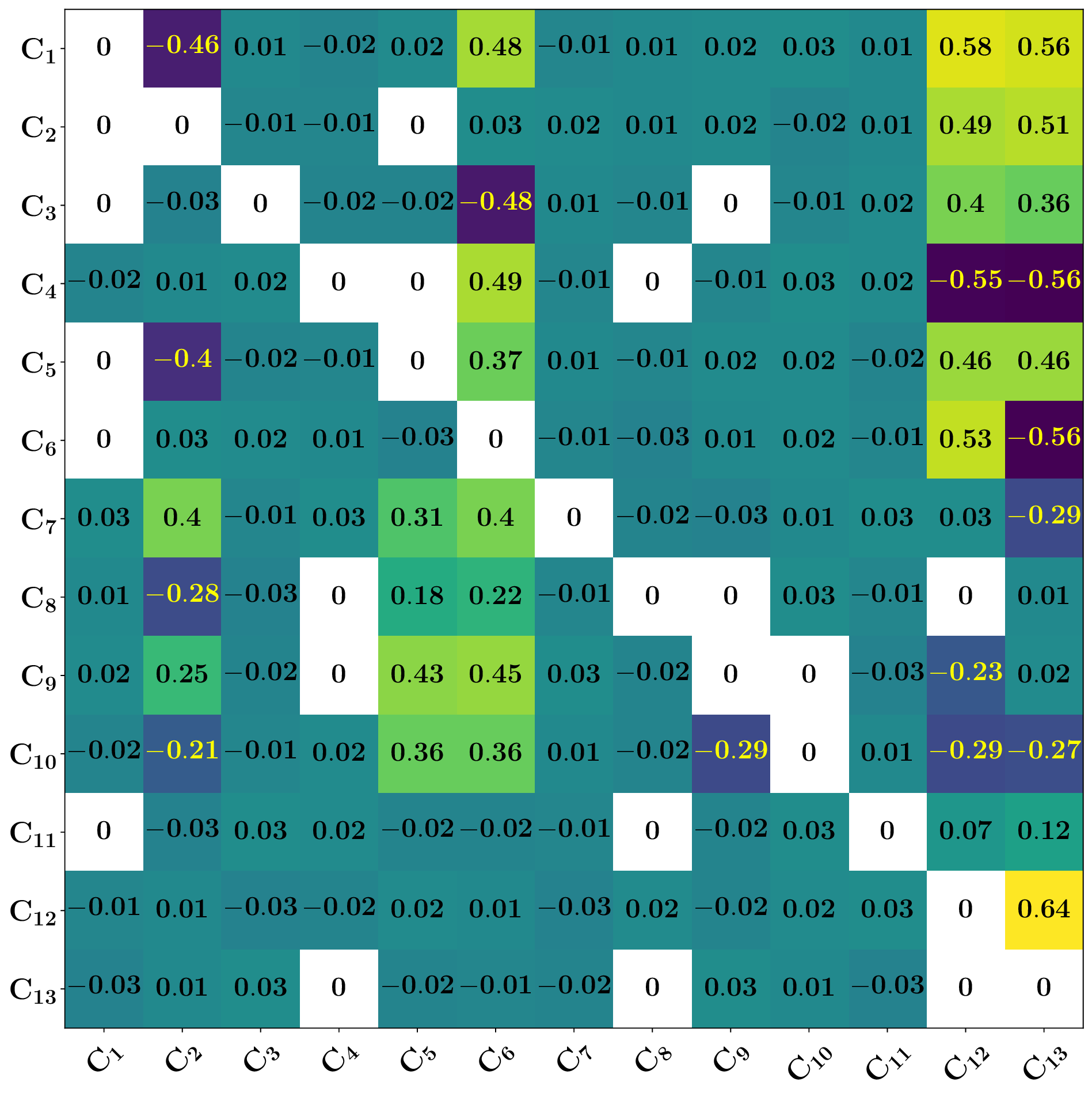} 
 	%ischemic_stroke_mixture_likelihood1_att1.eps
 	\caption{ Likelihood FCM $\mathcal{F}_{2}$}
    \label{fig:likelihood_2}
    \end{subfigure}
    \hfill
  \begin{subfigure}[t]{0.24\textwidth}
     \vspace{0.10in}
    \centering
 	\includegraphics[width=\linewidth]{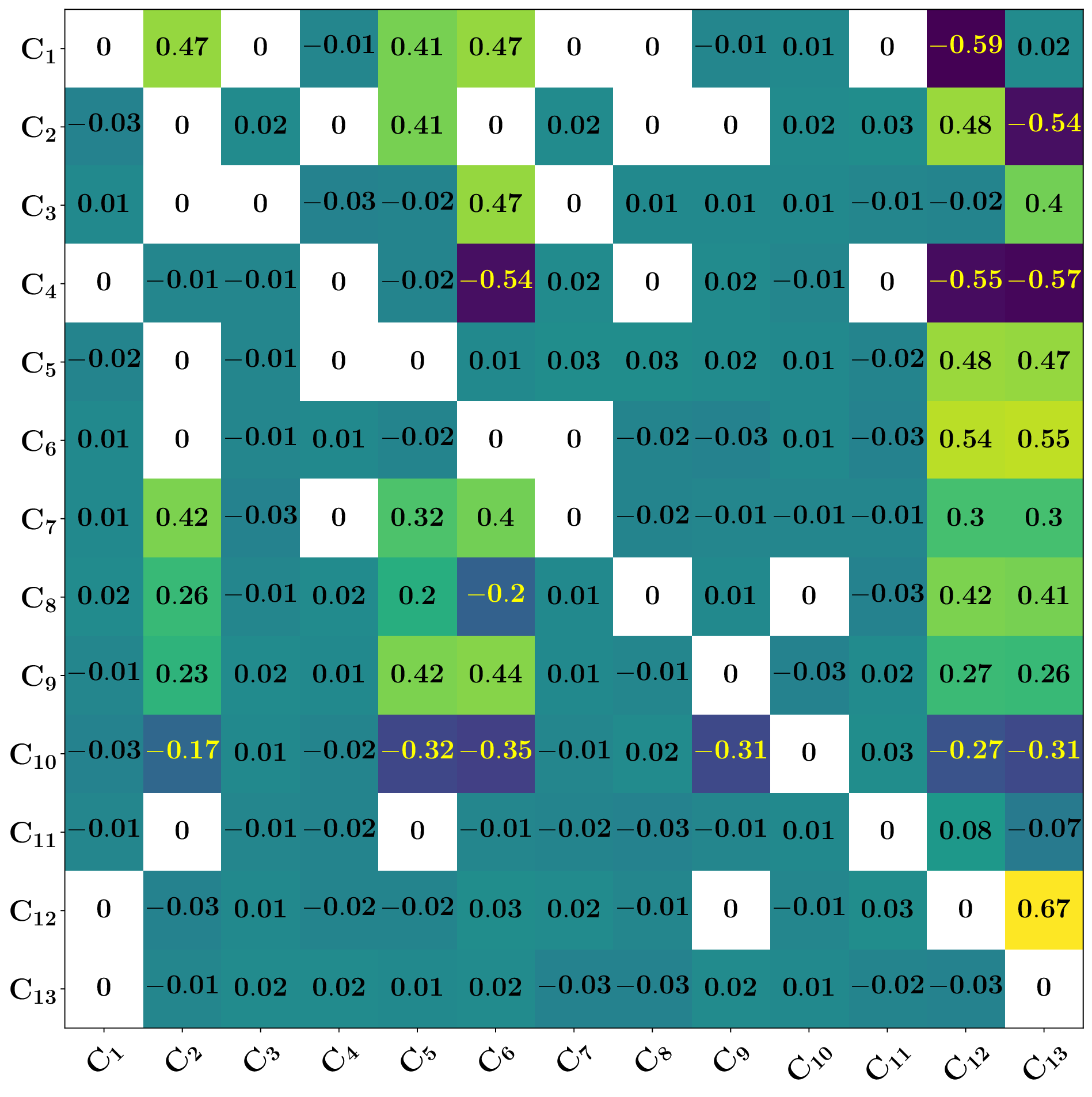}
 	\caption{ Likelihood FCM $\mathcal{F}_{3}$}
    \label{fig:likelihood_3}
    \end{subfigure}
    \hfill
  \begin{subfigure}[t]{0.24\textwidth}
     \vspace{0.10in}
    \centering
 	\includegraphics[width=\linewidth]{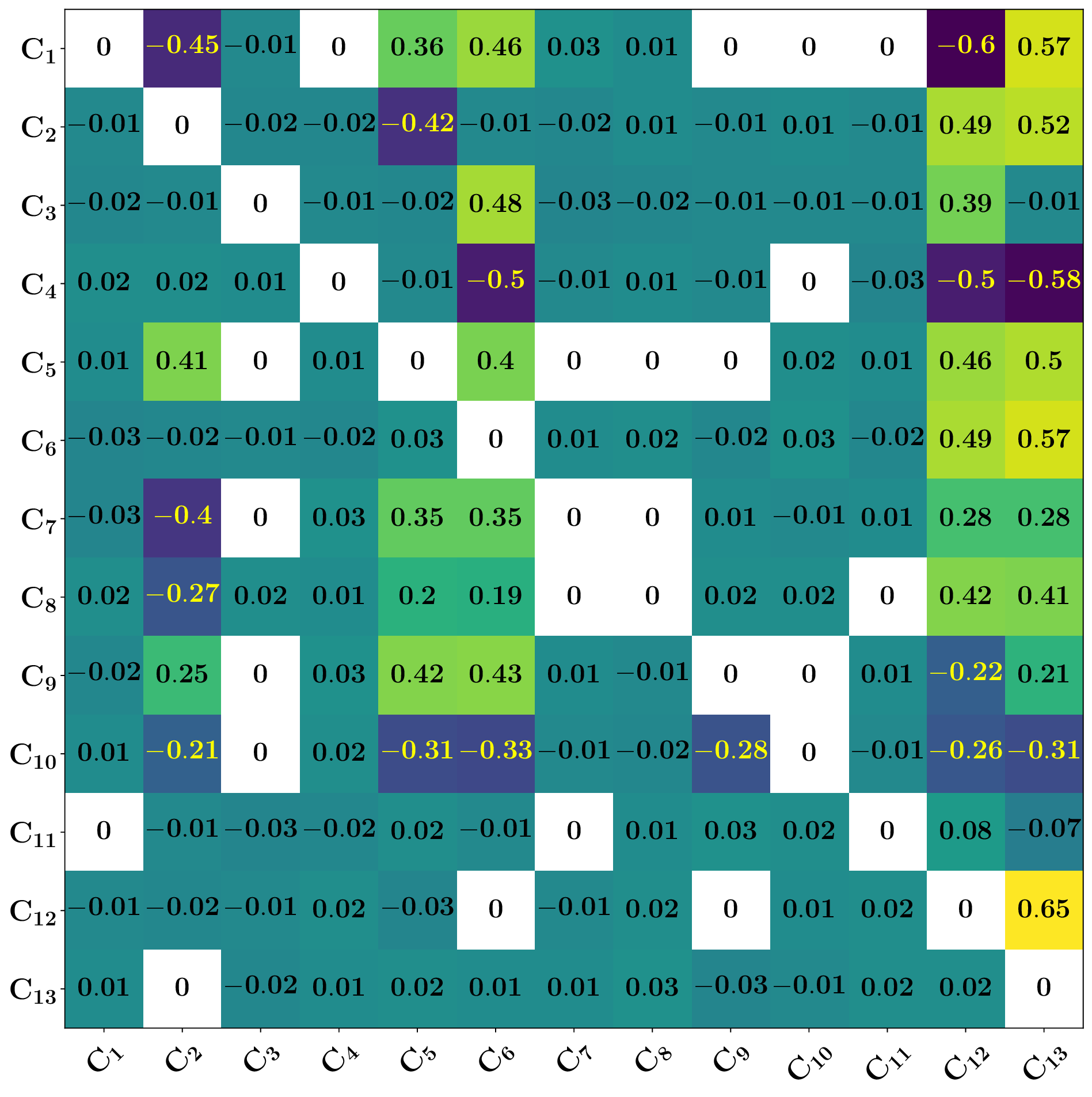}
 	\caption{ Likelihood FCM $\mathcal{F}_{4}$}
    \label{fig:likelihood_4}
    \end{subfigure}
  %\begin{comment}
  \begin{subfigure}[t]{0.24\textwidth}
     \vspace{0.10in}
   \centering
   \includegraphics[width=\linewidth]{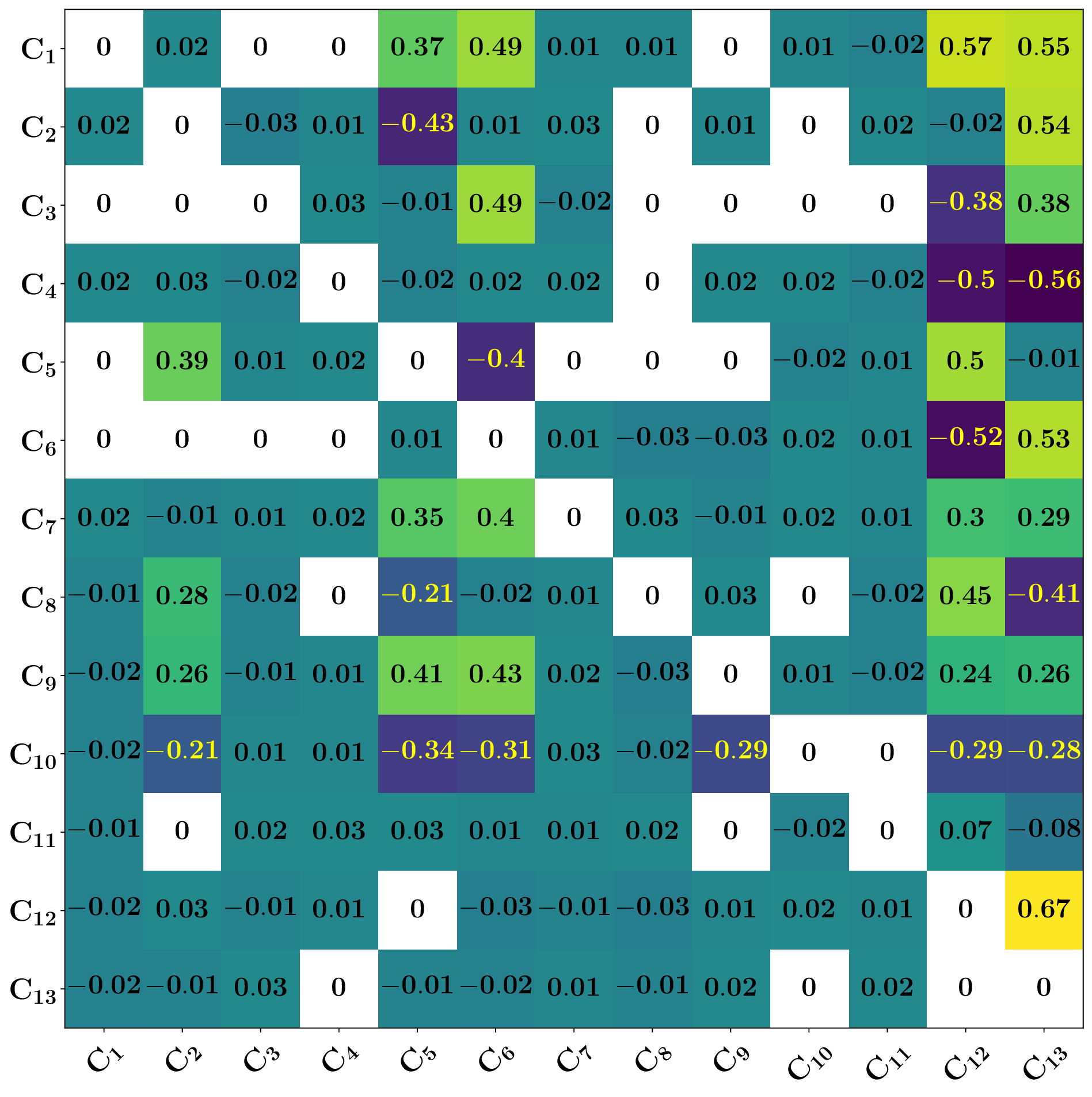}
   \caption{ Likelihood FCM $\mathcal{F}_{5}$}
   \label{fig:likelihood_5}
    \end{subfigure}
    \begin{subfigure}[t]{0.24\textwidth}
     \vspace{0.10in}
    \centering
   	\includegraphics[width=\linewidth]{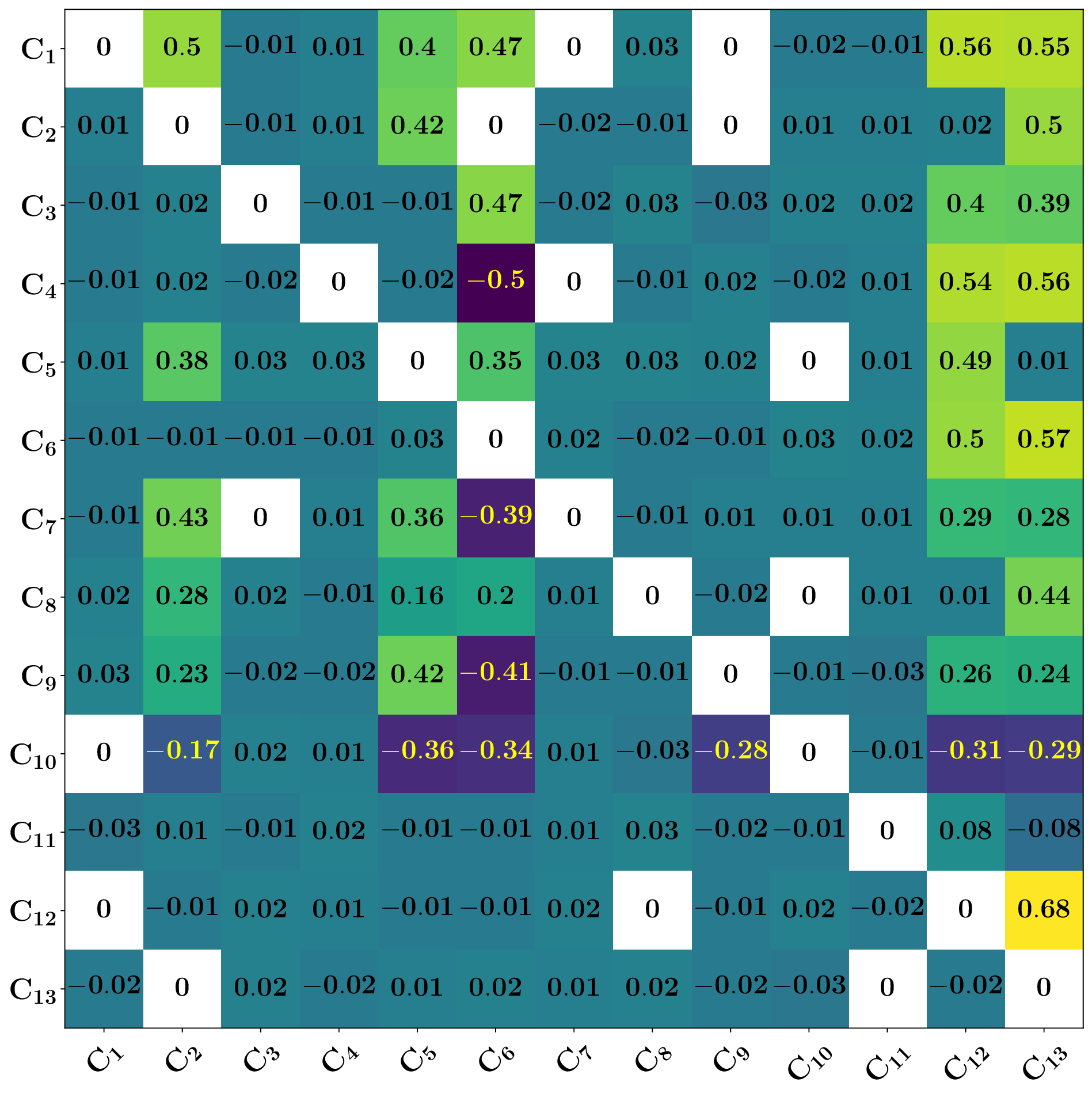}
   	\caption{ Likelihood FCM $\mathcal{F}_{6}$}
    \label{fig:likelihood_6}
    \end{subfigure}
   \begin{subfigure}[t]{0.24\textwidth}
    \vspace{0.10in}
    \centering
   	\includegraphics[width=\linewidth]{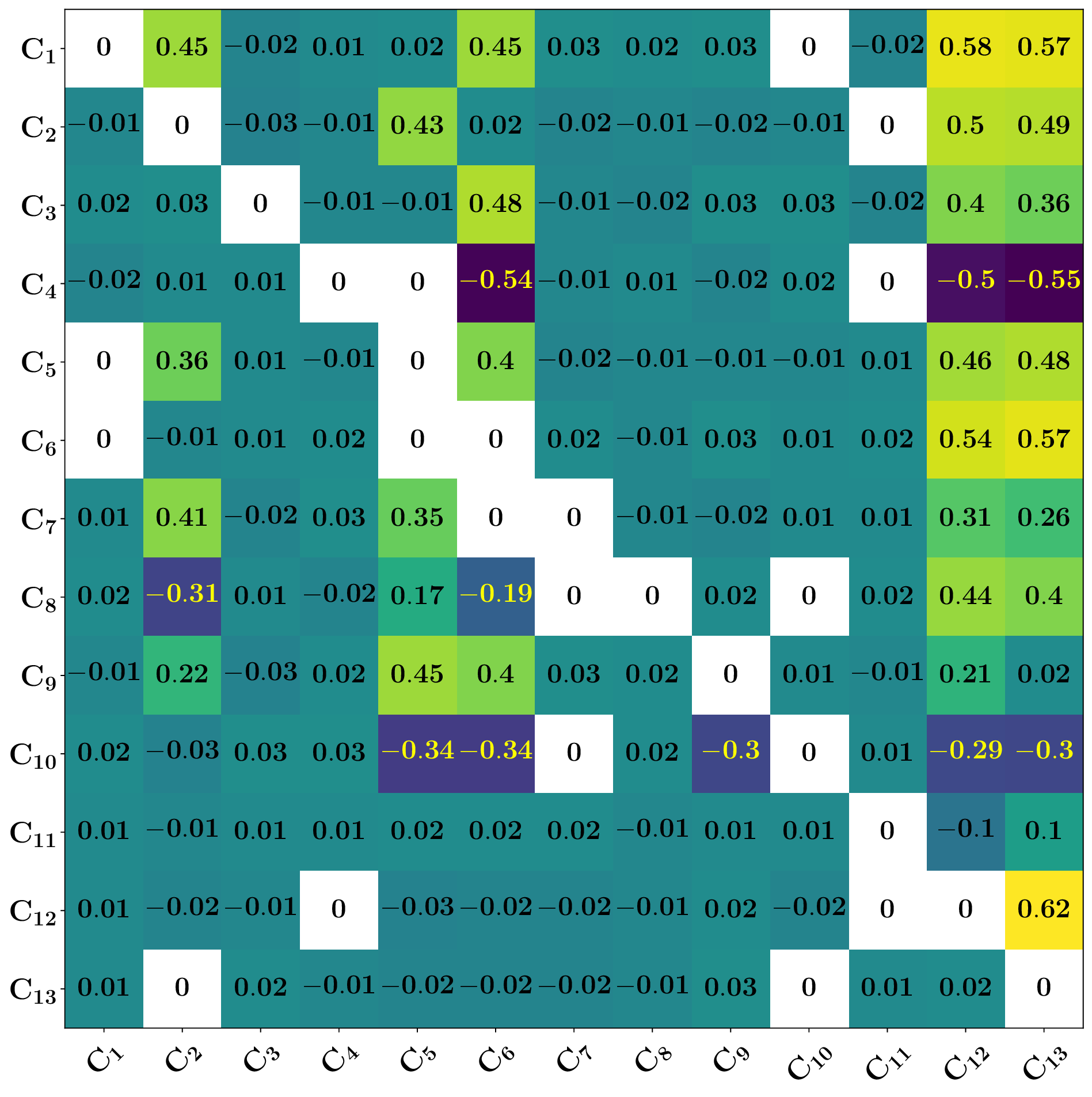}
   \caption{  Likelihood FCM $\mathcal{F}_{7}$}
    \label{fig:likelihood_7}
    \end{subfigure}
   %\end{comment}
   \caption{ \small Synthetic likelihood FCMs from the perturbation of the ischemic stroke FCM \cite{khodadadi2019fuzzy}: Edge matrices corresponding to the 7 likelihood FCMs.
   Brighter color corresponds to higher edge weight. 
   White corresponds to zero edge weight.
    }
    %\vspace{-0.0in}
 \end{figure*}

\begin{figure*}[htbp]
 \begin{subfigure}[t]{0.46\textwidth}
     \vspace{0.0in}
    \centering
 	\includegraphics[width=0.9\linewidth]{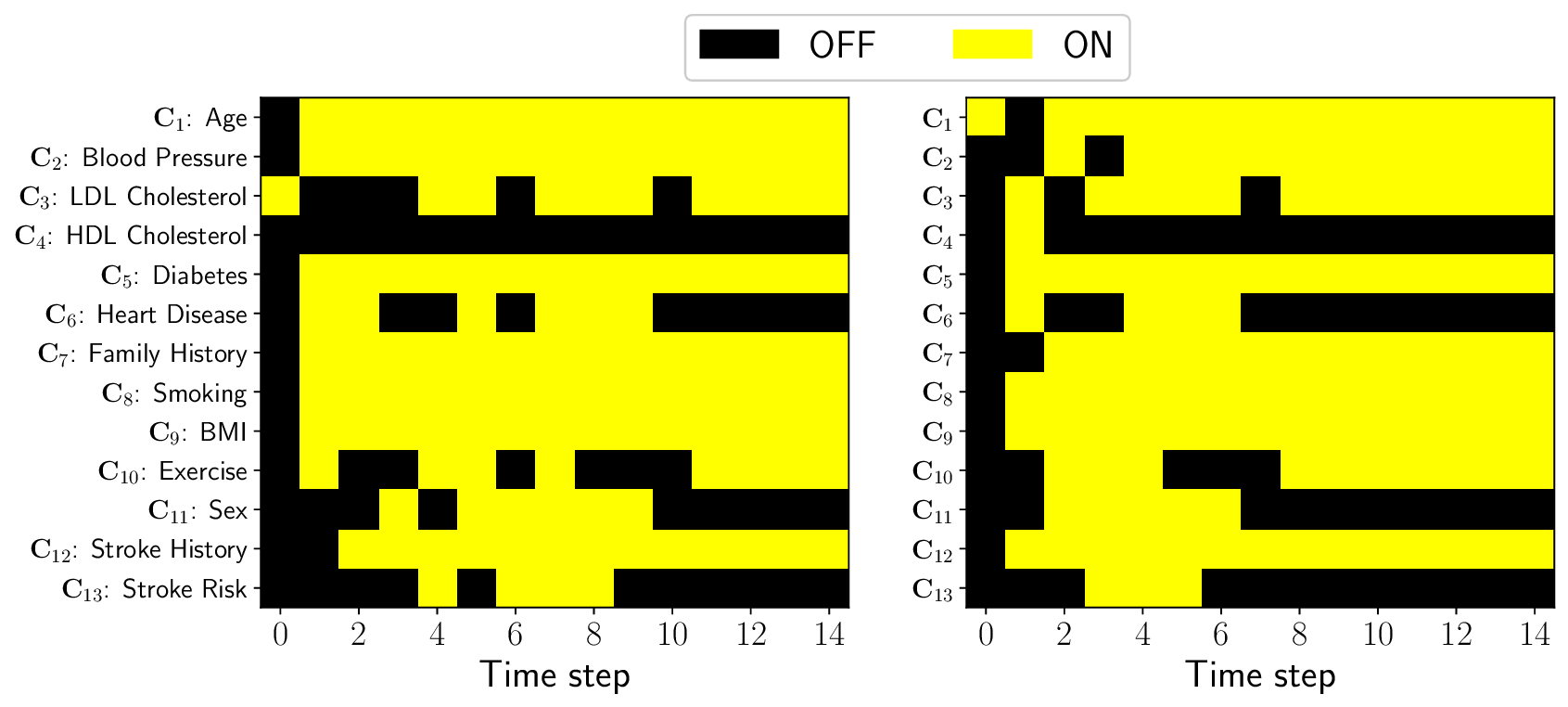}
 	\caption{$\mathcal{F}_1$'s equilibria}
    \label{fig:likelihood_1_attractor}
    \end{subfigure} 
     \begin{subfigure}[t]{0.53\textwidth}
     \vspace{0.0in}
    \centering
 	\includegraphics[width=0.9\linewidth]{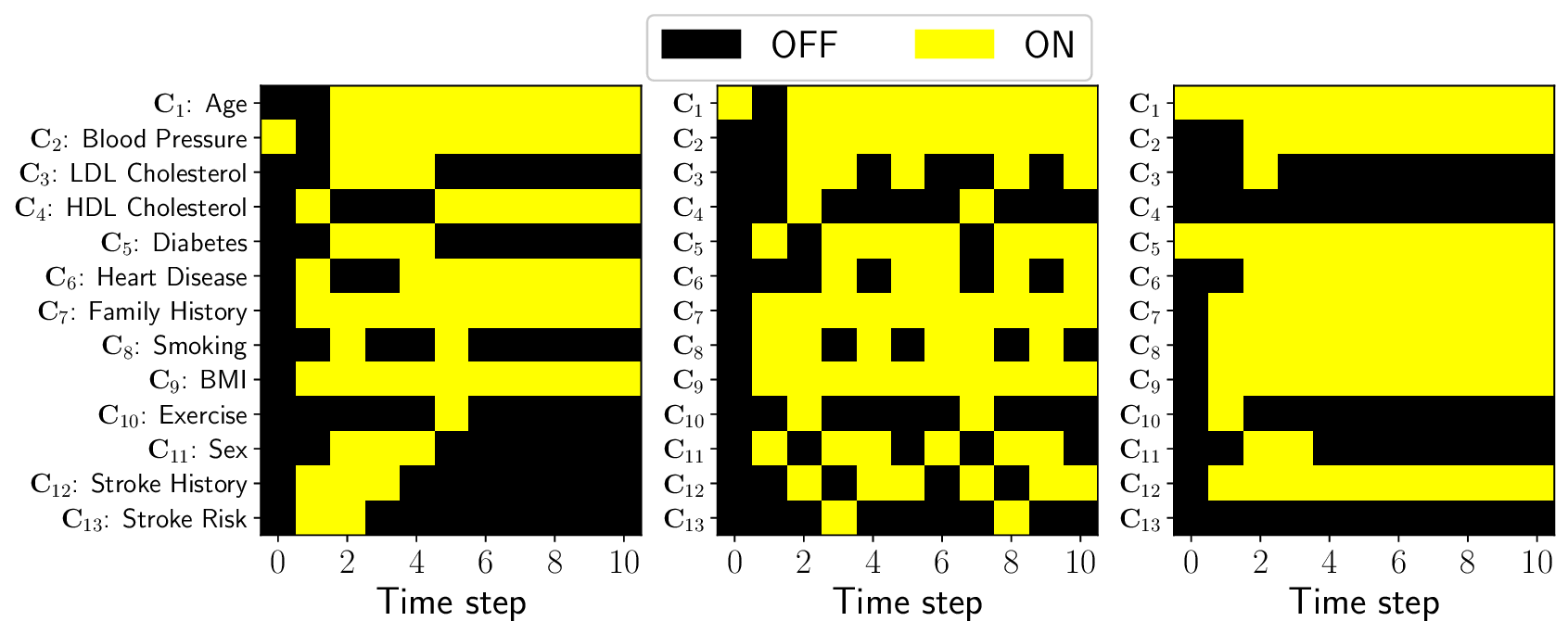}
 	\caption{$\mathcal{F}_2$'s equilibria}
    \label{fig:likelihood_2_attractor}
    \end{subfigure} 
    \begin{subfigure}[t]{0.53\textwidth}
     \vspace{0.05in}
    \centering
 	\includegraphics[width=0.9\linewidth]{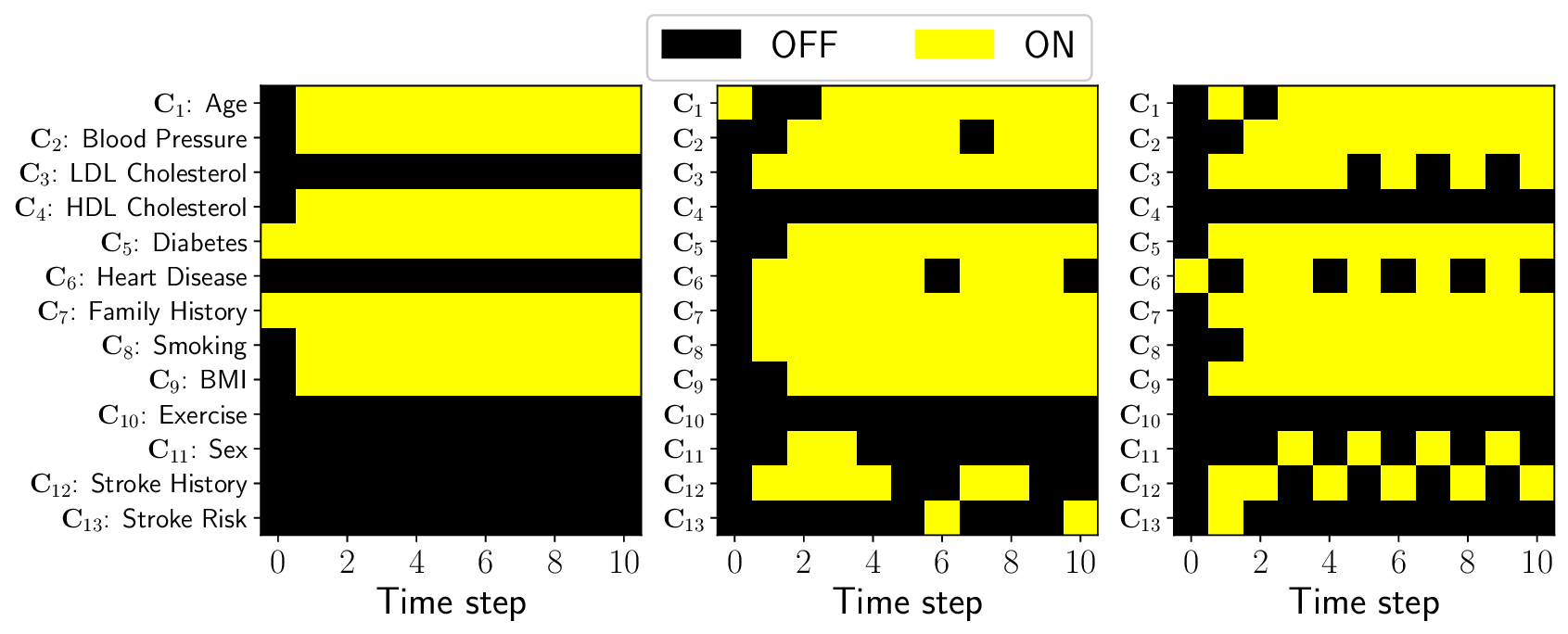}
 	\caption{$\mathcal{F}_3$'s equilibria}
    \label{fig:likelihood_3_attractor}
    \end{subfigure}
     \begin{subfigure}[t]{0.22\textwidth}
     \vspace{0.05in}
    \centering
 	\includegraphics[width=0.9\linewidth]{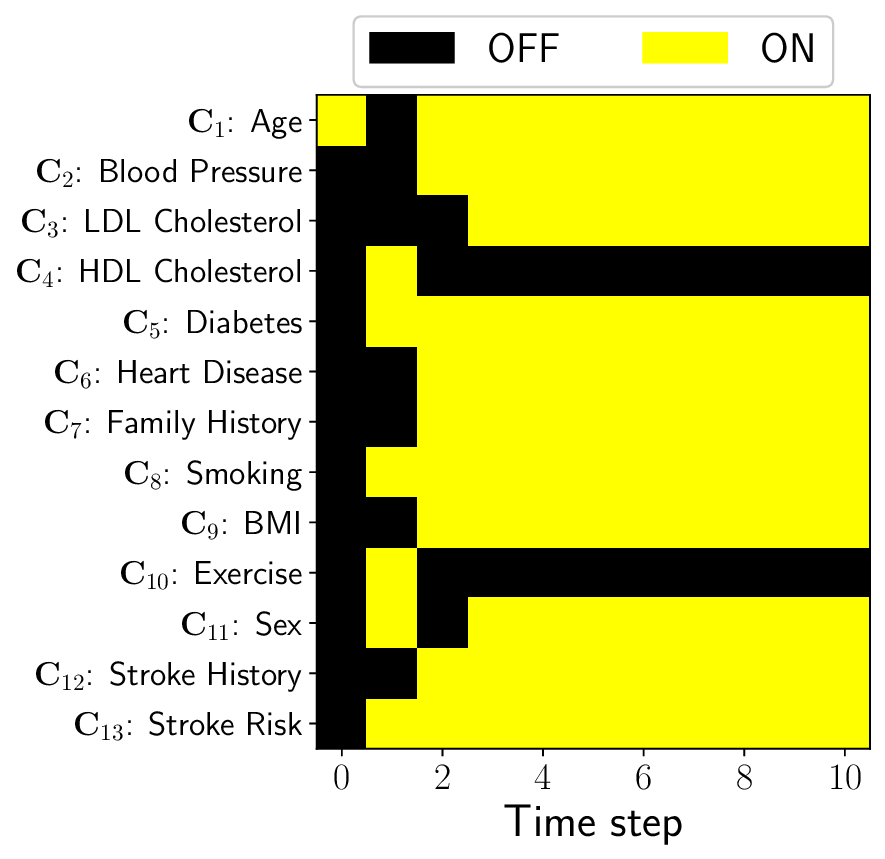}
 	\caption{$\mathcal{F}_4$'s equilibrium}
    \label{fig:likelihood_4_attractor}
    \end{subfigure}    
   \begin{subfigure}[t]{0.22\textwidth}
     \vspace{0.05in}
    \centering
 	\includegraphics[width=0.9\linewidth]{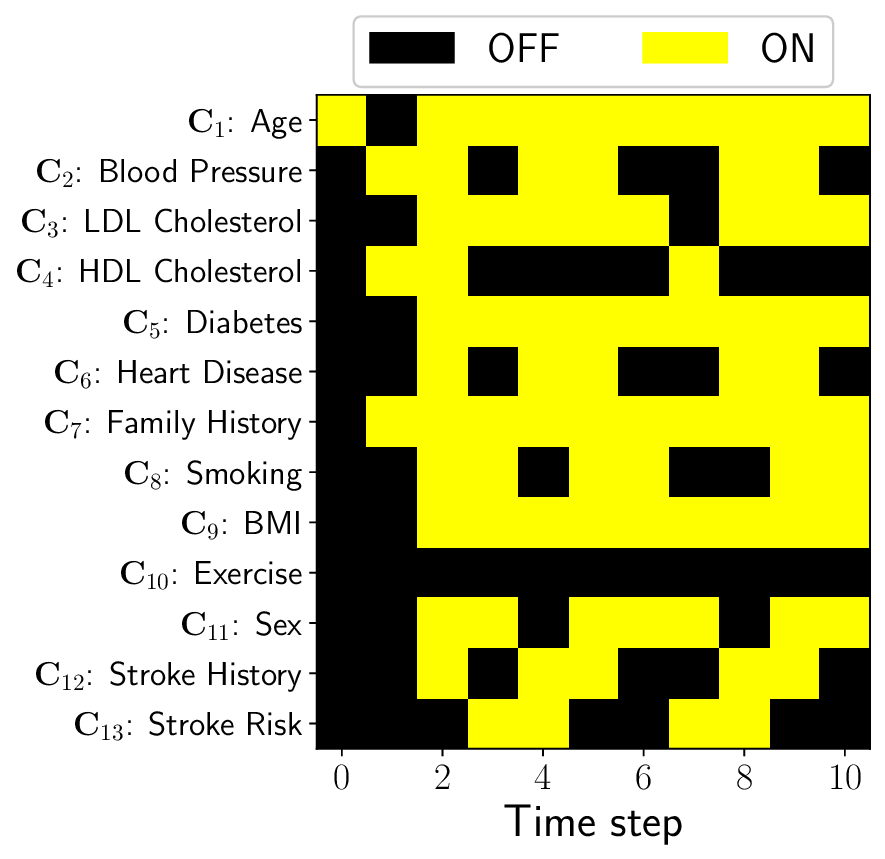}
 	\caption{$\mathcal{F}_5$'s equilibrium}
    \label{fig:likelihood_5_attractor}
    \end{subfigure} 
    %\begin{comment}  
     \begin{subfigure}[t]{0.50\textwidth}
     \vspace{0.05in}
    \centering
 	\includegraphics[width=0.9\linewidth]{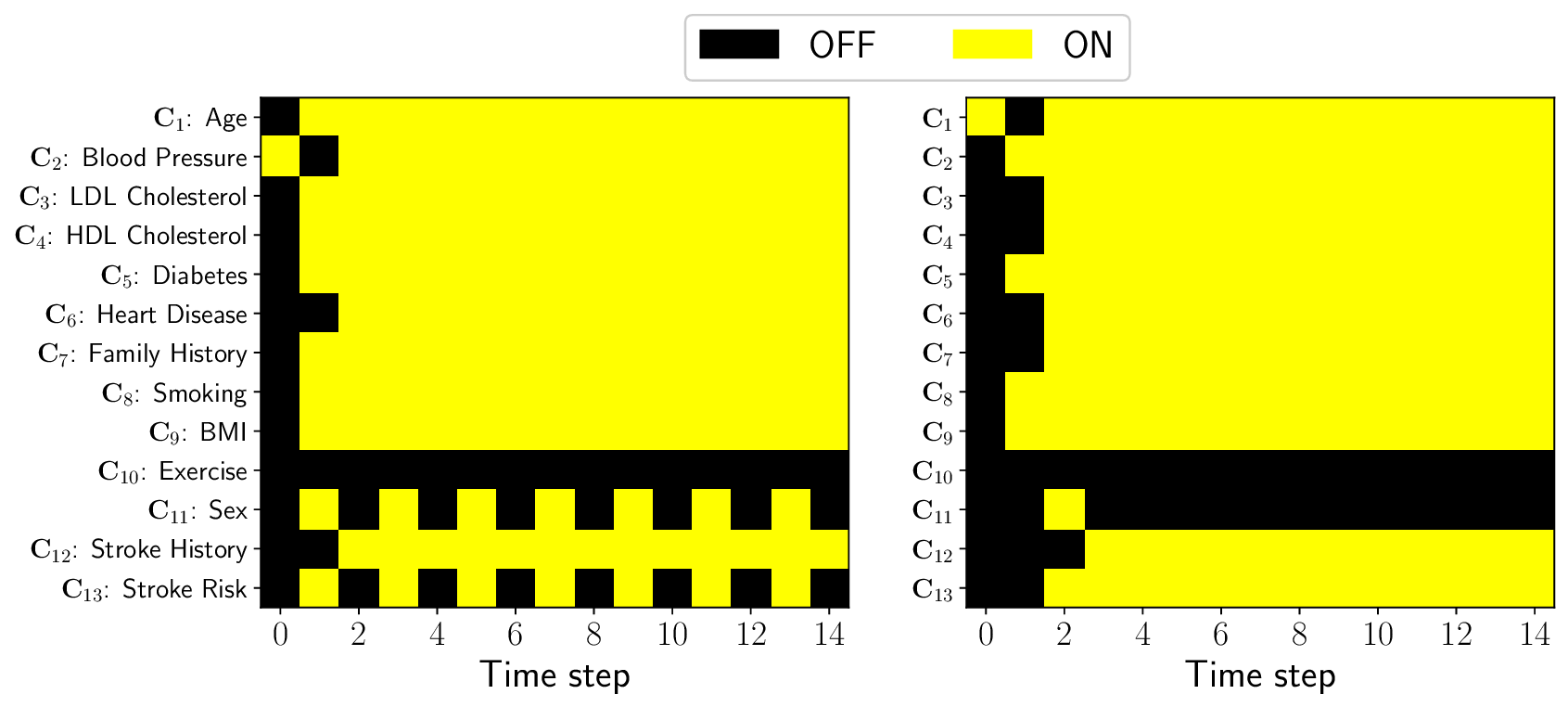}
 	\caption{$\mathcal{F}_6$'s equilibria}
    \label{fig:likelihood_6_attractor}
    \end{subfigure} 
    \begin{subfigure}[t]{0.50\textwidth}
     \vspace{0.05in}
    \centering
 	\includegraphics[width=0.9\linewidth]{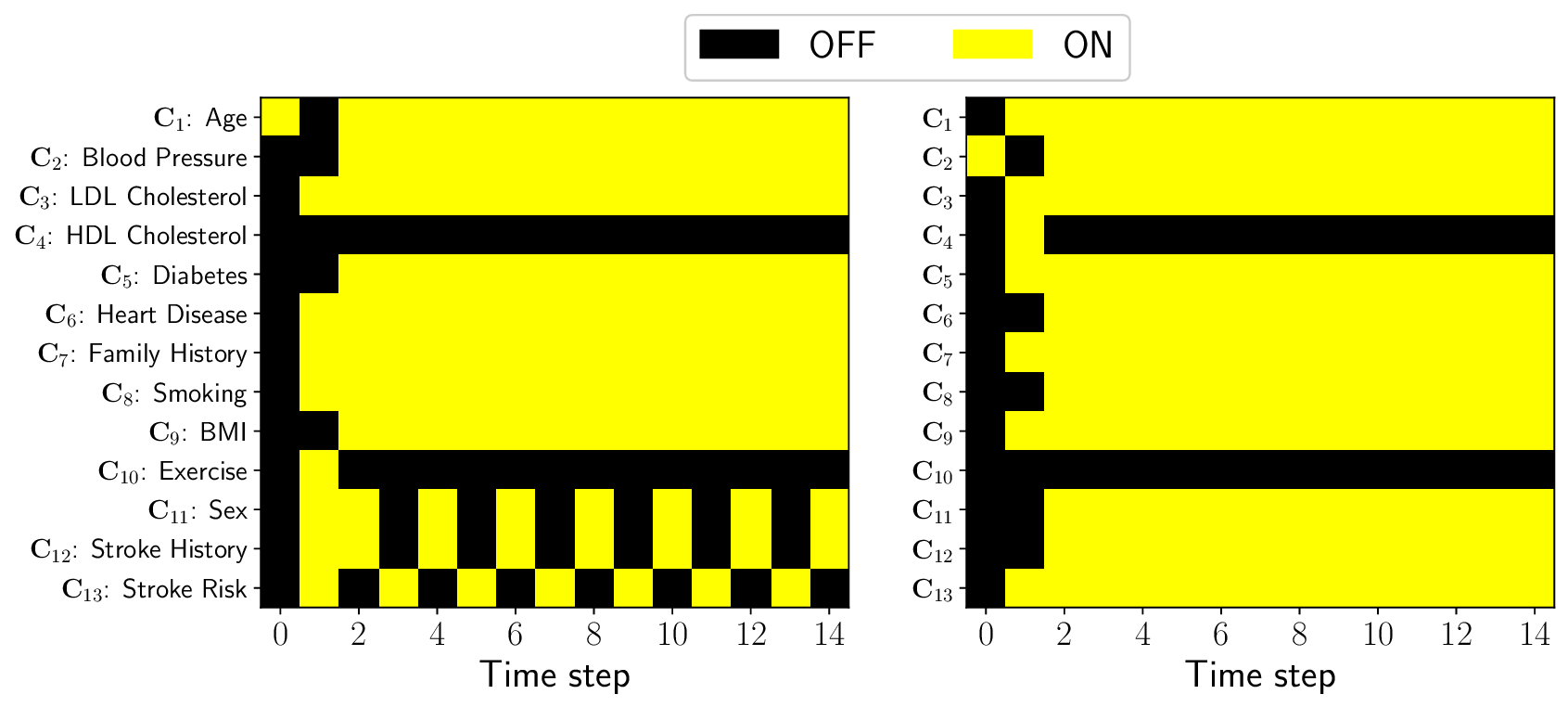}
 	\caption{$\mathcal{F}_7$'s equilibria}
    \label{fig:likelihood_7_attractor}
    \end{subfigure} 
%\end{comment}
       \caption{ \small Nontrivial attractors for the synthetic likelihood FCMs from the perturbation of the ischemic stroke FCM \cite{khodadadi2019fuzzy}: (a) $\mathcal{F}_1$ has two fixed points. (b) $\mathcal{F}_2$  has one limit cycle with length 5 and two fixed points. (c) $\mathcal{F}_3$  has  one fixed point, one limit cycle with length 4, and one limit cycle with length 2.  (d) $\mathcal{F}_4$  has just a fixed point. (e) $\mathcal{F}_5$ has a limit cycle with length 3.  
       (f)  $\mathcal{F}_6$ has a limit cycle with length 2 and a fixed point. 
       (g) $\mathcal{F}_7$ has a limit cycle with length 2 and one fixed point.
       The nodes are along the $y$-axis and the time steps are along the $x$-axis. 
       Yellow corresponds to active nodes and black corresponds to inactive nodes. %\vspace{-0.15in}
    }
 \end{figure*} 
 
  \begin{figure}[htbp]
\centering
   \begin{subfigure}[t]{0.48\textwidth}
    \centering
 	\includegraphics[width=0.75\linewidth]{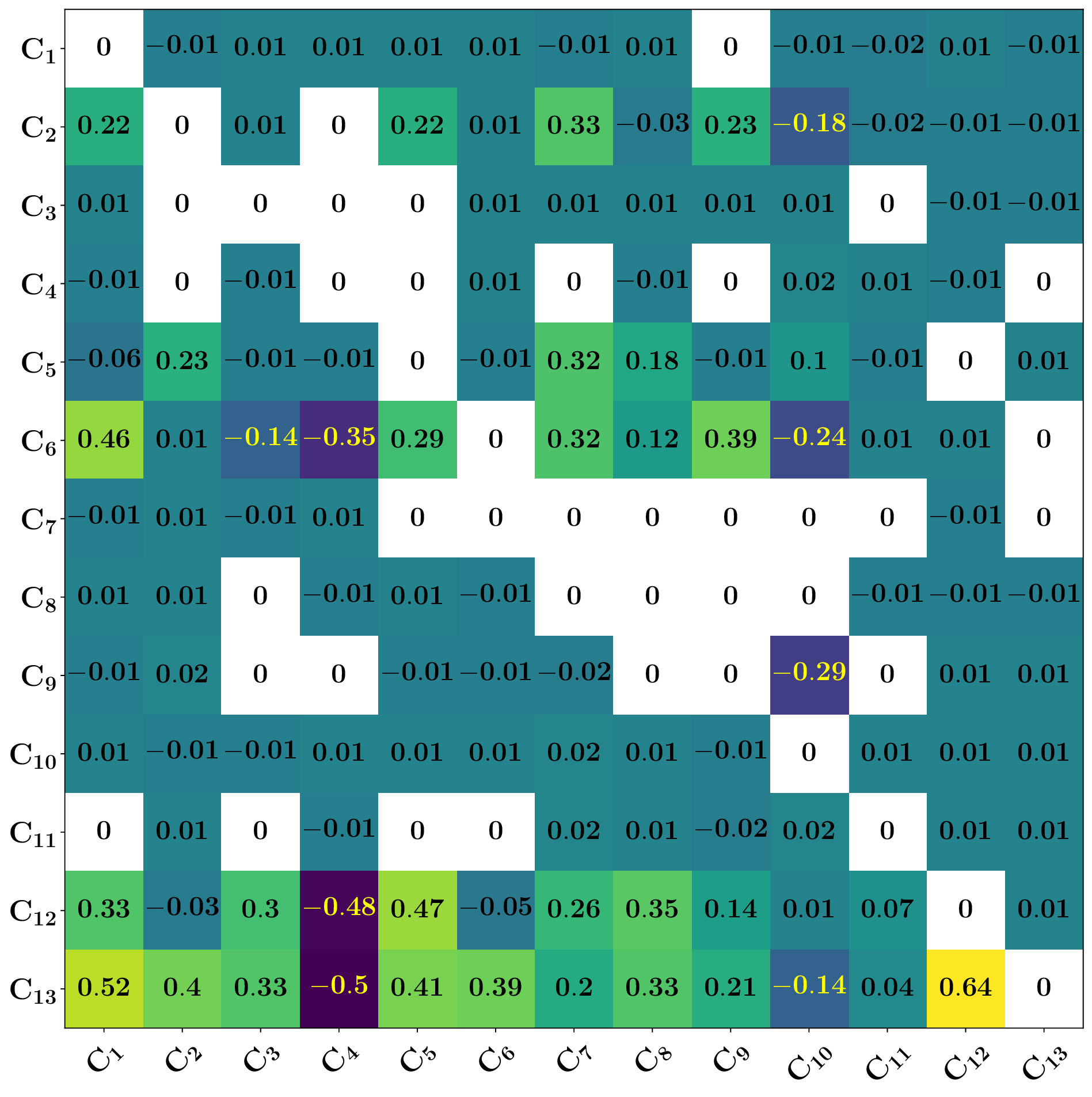}
 	\caption{Likelihood FCM mixture $\mathcal{F}$}
    \label{fig:likelihood_mix}
    \vspace{0.15in} 
    \end{subfigure}
   \begin{subfigure}[t]{0.48\textwidth}
    \centering
 	\includegraphics[width=0.75\linewidth]{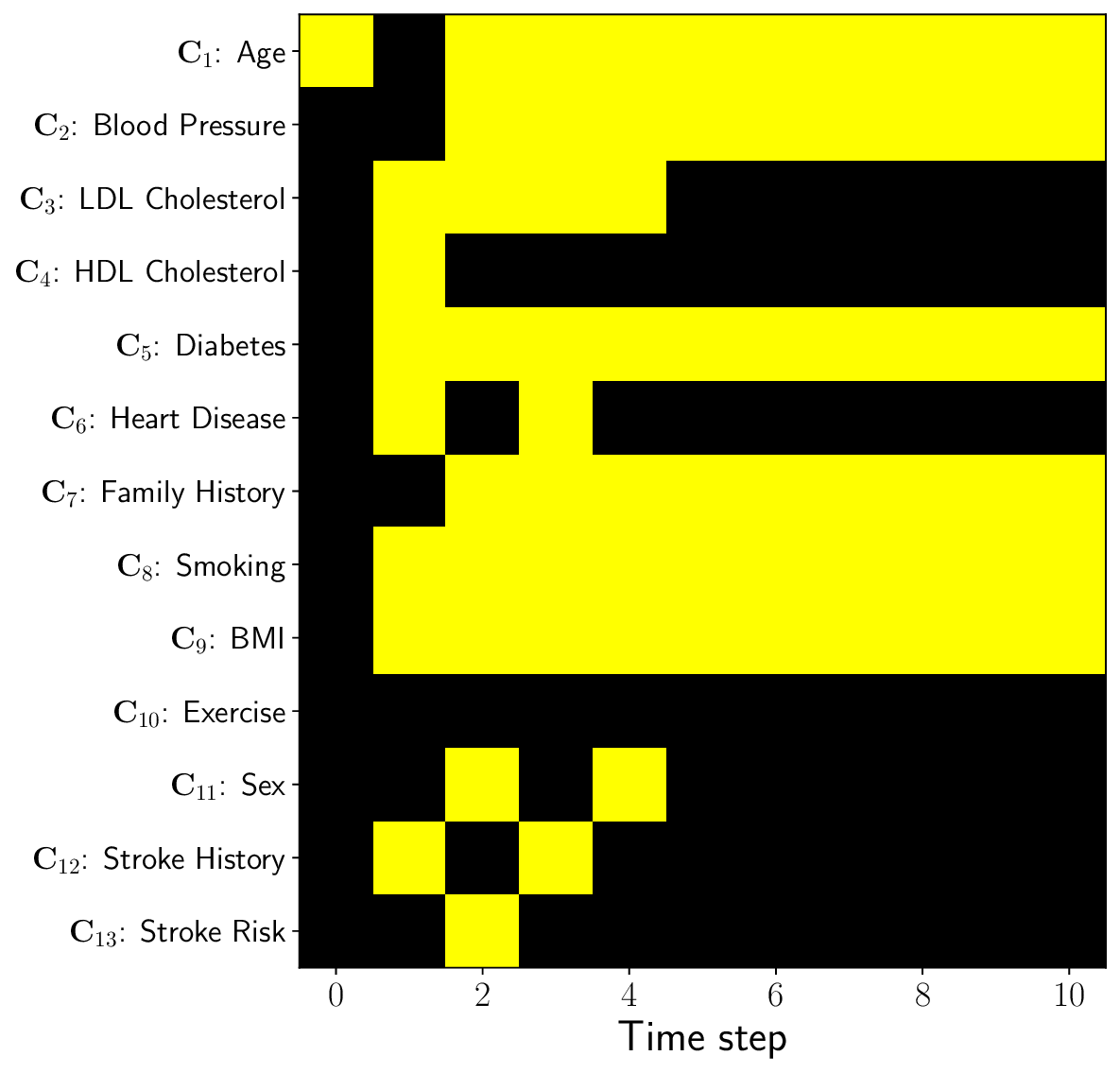}
 	\caption{ Attractor: One nontrivial fixed point}
    \label{fig:likelihood_mix_attractor}
    \end{subfigure}
   \caption{\small Mixing the likelihood FCMs with the set  of non-uniform weights  ($w_1= 0.5, w_2 = 0.15,  w_3= 0.1, w_4 = 0.1,  w_5 =  0.05, w_6 = 0.05, w_7 = 0.05$): 
   (a) The edge matrix for the non-uniformly mixed likelihood FCM.  (b) One of the nontrivial attractors in which blood pressure, diabetes, smoking, and BMI remain active.
    }
  %\vspace{-0.15in}
 \end{figure}
 
\subsection{Discrete-Time FCM Evolution}

An $n$-dimensional row vector $C(t) \in [0,1]^n$ describes the state of the FCM's concept nodes at time $t$. 
The $i^\text{th}$ node is ``active'' or ``on'' at time $t$ if the $i^\text{th}$ component $C_i(t)$ of the state vector $C(t)$ is equal to or close to one. 
The $i^\text{th}$ node is ``inactive'' or ``off'' at time $t$ if $C_i(t)$ is equal to or close to zero. 
A node is partially active otherwise. 
The causal variables corresponding to the active nodes are present in the system and those corresponding to the inactive nodes are absent. 
The causal factors are partially present in the system if their corresponding nodes are partially active. 

FCMs evolve in discrete time through vector-matrix multiplication and nonlinear squashing. 
The state $C_j(t+1)$ of the $j^\text{th}$ concept node $C_j$ at discrete time step $t+1$ is: %$C_j(t+1) = \Phi\Big(\sum_{i=1}^n  C_i(t)e_{ij}\Big)$
\begin{align}\label{eq:fcm-update}
   C_j(t+1) = \Phi\bigg(\sum_{i=1}^n  C_i(t)e_{ij}\bigg)
\end{align}
where $\Phi$ is a nonlinear function bounded between zero and one. 
The sum $\sum_{i=1}^n  C_i(t) e_{ij}$ is the $j^{\text{th}}$ component of the product between the state row-vector $C(t)$ and the edge matrix $E$. 
The nonlinear function $\Phi$ then squashes this product between zero and one. 
This process repeats itself to give the discrete-time evolution of the FCM. 
The FCM starts with the initial state $C(0)$ at time $t = 0$ and then goes through the states $C(1)$, $C(2)$, $C(3)$, and so on in order. 
The active nodes in this state-vector sequence qualitatively describe the trajectory of the dynamical system that the FCM models.

\subsection{FCM Dynamic Equilibria}

The equilibria are dynamic attractors that describe the causal what-if predictions of the FCM nonlinear dynamical system. 
The equilibrium behavior of the FCM depends on the limiting behavior of the state-vector sequence. 
The FCM converges to a \emph{fixed-point} attractor if the state-vector sequence converges to a constant vector. 
The FCM converges to a $K$-step \emph{limit cycle} for an integer $K > 1$ if the state-vector converges to a repeating cycle of states and so if $C(t+K) = C(t)$ holds somewhere in the state-vector sequence. 
So a fixed point is a trivial or 1-step limit cycle.
The FCM may also converge in general to a wandering chaotic attractor where there are no repeating patterns in the state-vector sequence. 

The set of all initial conditions $C(0)$ that lead to a given equilibrium describes the basin of attraction for that equilibrium. 
The FCM describes a map from these basins to their corresponding equilibrium attractors. 
The basins of the FCM's equilibria partition the FCM's input space. 
The FCM models a dynamical system by approximating its corresponding basin-to-equilibrium map. 

\subsection{FCM Mixtures via Zero-padding}

FCMs combine their knowledge through convex mixing. 
Consider $m$ FCMs. 
Say the $k^\text{th}$ FCM has the set of nodes $S_k$ and the edge matrix $E_k$. 
The node-set $S$ for the $n$-node FCM mixture is $S_1\cup S_2\cup S_3\cup ...\cup S_m$. 
The $n$-by-$n$ matrix $\Tilde{E_k}$ pads the $k^\text{th}$ edge matrix $E_k$ with zero rows and zero columns corresponding to the nodes in the set difference $S - S_k$. 
The edge matrix $E$ for the mixed FCM is
\begin{align}
    E = \sum_{k=1}^m w_k\Tilde{E_k} \label{eg:fcm_mixing}
\end{align}
where $w_k$ are convex mixing weights such that $w_k\geq0$ and $\sum_{k=1}^mw_k=1$. 
FCM mixing is closed.
This means that mixing FCMs gives back an FCM.
This is not always true for Directed Acyclic Graph (DAG) based causal models like Bayesian Belief Networks (BBNs) or for Markov chains.

\section{Bayesian-like Posterior FCMs}

\emph{Total probability} mixes likelihoods $P(E|H_j)$ using the prior probabilities $P(H_j)$ as convex mixing weights:  $P(E) = \sum_{j=1}^m P(H_j)P(E|H_j)$ for evidence $E$ and disjoint exhaustive hypotheses $H_j$. 
Then Bayes theorem gives the  posterior probabilities $P(H_k|E)$:  $P(H_k|E)=P(H_k)P(E|H_k)/P(E)$ that further simplifies to  $P(H_k|E)=P(H_k)P(E|H_k)/\sum_{j=1}^m P(H_j)P(E|H_j)$ because of total probability. 

Equation~(\ref{eg:fcm_mixing}) shows that mixed FCMs also share the same convex mixing structure as does total probability. 
But they mix edge \emph{matrices} instead of mixing likelihood probabilities.
This gives a new Bayesian-like posterior FCM \cite{panda2026thucydides}. 
Say $m$ likelihood FCMs $\mathcal{F}_1$, $\mathcal{F}_2$, ... , $\mathcal{F}_m$ with edge matrices $E^l_1$, $E^l_2$, ... , $E^l_m$ mix with respective mixing weights $w_1$, $w_2$, ... , $w_m$. 
Then the posterior FCM $\mathcal{P}_k$ that corresponds to the $k^\text{th}$ FCM $\mathcal{F}_k$ has the edge matrix $E^p_k$: 
\begin{align}\label{eqn:4}
    E^p_k &= w_k\tilde{E}^l_k(E^l)^{-1} = w_k\tilde{E}^l_k\Bigg(\sum_{j=1}^mw_j\tilde{E}^l_j\Bigg)^{-1}
\end{align}
if the mixed FCM's edge matrix $E^l=\sum_{j=1}^mw_j\tilde{E}^l_j$ is invertible. If $E^l$ is not invertible then we use the \emph{pseudo-inverse} of $E^l$:
\begin{align}
 E^p_k &= w_k\tilde{E}^l_k(E^l)^{+}
 \label{eq:posterior_FCM}
\end{align}
where $(E^l)^{+} = (E^l)^T(E^l(E^l)^T)^{-1}$ if $E^l$ is a full row matrix and  $(E^l)^{+} = ((E^l)^TE^l)^{-1} (E^l)^T$ if $E^l$ is a full column matrix.  
The singular value decompistion approach works for the pseudo-inverse if $E^l$ is neither full column rank or full row rank. 
%\begin{align} \label{eq:posterior_FCM}
 %   E^p_k &= w_k\tilde{E}^l_k(E^l)^T(E^l(E^l)^T)^{-1} . % \\
    %&= w_k\tilde{E}^l_k\Bigg(\sum_{j=1}^mw_j\tilde{E}^l_j\Bigg)^T\Bigg(\Bigg(\sum_{j=1}^mw_j\tilde{E}^l_j\Bigg)\Bigg(\sum_{j=1}^mw_j\tilde{E}^l_j\Bigg)^T\Bigg)^{-1}  . \label{eq:posterior_FCM}
%\end{align}

The posterior FCM may have edge weights outside the bipolar interval $[-1,1]$. 
We can set the magnitude of those edge weights to equal 1. 
%\begin{figure}[htbp]
%\centering
%   \begin{subfigure}[t]{0.40\textwidth}
%    \centering
% 	\includegraphics[width=0.9\linewidth]{Figures/ischemic_weight_likelihood_non_uniform_mixture_2.eps}
% 	\caption{Likelihood FCM mixture $\mathcal{F}$}
 %   \label{fig:likelihood_mix2}
 %   \end{subfigure}
 %   \hfill 
 %  \begin{subfigure}[t]{0.4\textwidth}
 %   \centering
% 	\includegraphics[width=\linewidth]{Figures/ischemic_weight_likelihood_non_uniform_mixture_2_att.eps}
% 	\caption{ Attractor: One nontrivial fixed point}
  %  \label{fig:likelihood_mix2_attractor}
 %   \end{subfigure}
 %  \caption{ Mixing the likelihood FCMs with a set  of non-uniform weights  ($w_1= 0.12, w_2 = 0.07,  w_3= 0.18, w_4 = 0.05,  w_5 =  0.23, w_6 = 0.14, w_7 = 0.21$)
 %   }
% \end{figure}
Posterior FCMs may not have the same nodes or edges as their corresponding likelihood FCMs. 
The posterior FCMs' edge matrices $E^p_k$ add to the identity matrix $\bf{I}$ just as the discrete posterior probabilities would add up to 1.
%\begin{align}
%    \sum_{k=1}^m E^p_k  = \bf{I} \label{sums-to-I} .
%\end{align}
The posterior FCMs then also mix to give the posterior-FCM mixture $\mathcal{P}_{\textsf{avg}}$. %: $E^p=\sum_{k=1}^m v_kE^p_k$
%\begin{align}
%   E^p=\sum_{k=1}^m v_kE^p_k \label{eq:posteiror-mixing}
%\end{align}
%where $v_k$ are convex mixing weights: $v_k \geq 0$ and $\sum_{k=1}^mv_k=1$. 

\section{FCM Mixture Relevance}

The Bayesian-like posterior FCM $\mathcal{P}_k$ updates the causal knowledge from the likelihood FCM $\mathcal{F}_k$ based on the information in the rest of the FCM mixture $\mathcal{F}$. 
Such posterior FCMs also combine with the same prior mixing weights $w_k$ as the likelihood FCMs to give the posterior FCM mixture $\mathcal{P}$. 

Let the $j^\text{th}$ posterior FCM $\mathcal{P}_j$ converge to the $K_j$-step limit cycle $\mathcal{L}_j$ for some initial state $x \in \{0,1\}^n$. 
Let the posterior FCM mixture $\mathcal{P}$ converge to a $K_p$-step limit cycle $\mathcal{L}$ for the same initial state $x$. 
Let $K = \text{max}(K_j, K_p)$ where both are at least one. 

Consider the $K$ consecutive state vectors $C^{(j)}(t_j)$, $C^{(j)}(t_j+1)$, ..., $C^{(j)}(t_j+K-1)$ from posterior FCM $\mathcal{P}_j$'s equilibrium $\mathcal{L}_j$ and compare them to the $K$ consecutive state vectors $C^{(p)}(t_p)$, $C^{(p)}(t_p+1)$, ..., $C^{(p)}(t_p+K-1)$ from the posterior FCM mixture $\mathcal{P}$'s equilibrium $\mathcal{L}$. 
The norm 
\begin{align}
    \varepsilon_j(x) = \Bigg(\frac{1}{K}\sum_{k=0}^{K-1}||C^{(j)}(t_j+k) - C^{(p)}(t_p+k)||^2_2\Bigg)^\frac{1}{2} \label{eq:compare-equilibria}
\end{align}
measures the distance between the equilibria and is high when the posterior FCM $\mathcal{P}_j$ predicts something significantly different from the posterior FCM mixture $\mathcal{P}_{\textsf{avg}}$. 
The time steps $t_j$ and $t_p$ can vary to maximize this distance $\varepsilon_j(x)$. 
The likelihood FCMs corresponding to such posterior FCMs are more relevant because their information is more probative. 
These distances $\varepsilon_j(x)$ then give the relevancy mixing weights:% $ r_j(x) = \frac{\varepsilon_j(x)}{\sum_{k=1}^m\varepsilon_k(x)} $.
\begin{align}
    r_j(x) = \frac{\varepsilon_j(x)}{\sum_{k=1}^m\varepsilon_k(x)}. \label{eq:relevant-weight}
 \end{align}
The likelihood FCMs $\mathcal{F}_j$ then mix using the relevancy mixing weights to give the relevant FCM mixture $\mathcal{F}_R(x)$ with the edge matrix 
%$E^R(x) = \sum_{j=1}^m r_j(x)\tilde{E}^l_j $.
\begin{align}
    E^R(x) = \sum_{j=1}^m r_j(x)\tilde{E}^l_j . \label{eq:relevant-mixture}
\end{align}

\section{Generating Synthetic FCM Experts}

Perturbations generate synthetic expert FCMs from the stroke FCM \cite{khodadadi2019fuzzy}. 
The FCMs $\mathcal{F}_k$ act as likelihood FCMs.  \\

{\it Generating Likelihood FCMs}: 
We jointly apply multiplicative and additive perturbations to the ischemic stroke  FCM \cite{khodadadi2019fuzzy} to generate likelihood FCMs.
Let ${\bf E} \in [-1, 1]^{n \times n} $ denote the connection weight matrix for the ischemic stroke FCM. 
The  likelihood FCM generator $\bf{F}$  from $\bf{E}$ is
\begin{align}\label{eq:noise_pertubation}
\bf{F} &= c (\bf{E} \circ  \bf{T}) + {\it d}(\bf{U}  \circ \bf{M})
\end{align}
where $\bf{T}$ is a random trivalent mask with $T_{ij} \sim$ Multinoulli$ (\{-1,0,1 \} )$, ${\bf U}$ is a random matrix with $U_{ij} \sim $ Uniform$([-1,1])$, $c \geq 0$, and $d \geq 0$.
${\bf M}$ represents the mask matrix for the additive perturbation: ${\bf M} = \mathbf{1}_n - \mathbf{I}_n$,  
%\begin{align}
%{\bf M} = \mathbf{1}_n - \mathbf{I}_n
%\end{align} 
where $\mathbf{1}_n$ is a matrix of all 1's and $\mathbf{I}_n$ is an identity matrix. 
The probability mass function for the multinoulli random variable $T_{ij}$ is 
\begin{align}\label{eq:multinouli_mask}
P(T_{ij} = t)  = 
\begin{cases}
p, \  & \text{if}  \ \   t = -1, \\
q,  \  & \text{if}  \ \   t = 0, \\
1 - p -q,  \  & \text{if}   \ \  t = 1, \\
\end{cases}
\end{align}
where $p,q \in [0,1]$.
%Therefore, the expectation  $\mathbb{E}[{\bf F}]$ of the likelihood generator  is $ c(1-2p-q){\bf E}$. 
The likelihood FCMs $\mathcal{F}_k$ mix through Equation~(\ref{eg:fcm_mixing}) with convex mixing coefficients $w_k$ that reflect our confidence in those experts. 
These mixing weights act as priors.

The mixture $\mathcal{F}$ gives Bayesian posterior-like FCMs $\mathcal{P}_k$ using Equation~(\ref{eq:posterior_FCM}) and (\ref{eq:posterior_FCM}). 
The posterior FCMs then also mix with the same weights $w_k$ to give the posterior mixture $\mathcal{P}$. 
The posterior FCMs and their mixture produce equilibria that give the relevancy weights $r_k(x)$ for given input $x$ using Equations~(\ref{eq:compare-equilibria}) and (\ref{eq:relevant-weight}). 
The likelihood FCMs $\mathcal{F}_k$ then mix using the relevance weights $r_k(x)$ to give the relevant FCM mixture $\mathcal{F}_R(x)$.

%\begin{figure*}[htbp]
%\centering
%   \begin{subfigure}[t]{0.40\textwidth}
%    \centering
% 	\includegraphics[width=0.9\linewidth]{Figures/ischemic_weight_mix_likelihood_nonuniform_relevance_2.eps}
% 	\caption{Relevant FCM mixture}
%    \label{fig:likelihood_mix2}
%    \end{subfigure}
%    \hfill 
%   \begin{subfigure}[t]{0.4\textwidth}
%    \centering
% 	\includegraphics[width=\linewidth]{Figures/ischemic_weight_mix_likelihood_nonuniform_relevance_att2.eps}
% 	\caption{ Attractor for  initial state}
%    \label{fig:likelihood_mix2_attractor}
%    \end{subfigure}
%   \caption{Relevant FCM mixture from the weighted likelihood mixture with the \emph{non uniform weights} and initial state $x^{(1)} = [0,0,0,0,1,1,0,0,0,0,0,0,0]$: 
%    }
%    \label{fig:relevant_fcm_mixture_with_nonuniform_wgts_2}
% \end{figure*}

 \begin{table}[htbp]
\caption{\small Relevancy mixing weights for the posterior FCMs with the non-uniform weights  ($w_1= 0.5, w_2 = 0.15,  w_3= 0.1, w_4 = 0.1,  w_5 =  0.05, w_6 = 0.05, w_7 = 0.05$). We picked four initial states $x^{(i)}$'s from the four largest basins.  }
\centering
\footnotesize
\begin{tabular}{|c|c|c|c|c|}
\hline
\multirow{2}{*}{Posterior $\mathcal{P}_j$} & \multirow{2}{*}{$r_j\big{(}x^{(1)}\big{)}$} & \multirow{2}{*}{$r_j\big{(}x^{(2)}\big{)}$} & \multirow{2}{*}{$r_j\big{(}x^{(3)}\big{)}$} & \multirow{2}{*}{$r_j\big{(}x^{(4)}\big{)}$} \\
&   &  &  & \\
\hline
$\mathcal{P}_1$                       & $0.083$  &  $0.127$  & $0.097$  & $0.083$  \\
$\mathcal{P}_2$           & $0.132$ &  $0.142$ & $0.149$  &  $0.156$ \\
$\mathcal{P}_3$                 & \color{blue}{$\bf{0.177}$}  &  $0.156$ &  $0.149$ &  \color{blue}{$\bf{0.177}$}  \\
$\mathcal{P}_4$         & $0.144$ & $0.142$ &  $0.149$ &  $0.118$ \\
$\mathcal{P}_5$   & \color{blue}{$\bf{0.177}$} &  \color{blue}{$\bf{0.180}$}  &  \color{blue}{$\bf{0.159}$}  & \color{blue}{$\bf{0.177}$}  \\
$\mathcal{P}_6$           & $0.132$ & $0.127$ &  $0.138$ &  $0.156$ \\
$\mathcal{P}_7$           & $0.156$ & $0.127$  & \color{blue}{$\bf{0.159}$} &  $0.132$ \\
\hline
\end{tabular}
\label{tab:table_2}
%\vspace{-0.15in}
\end{table}

\begin{figure*}[htbp]
   \begin{subfigure}[t]{0.27\textwidth}
    \centering
 	\includegraphics[width=\linewidth]{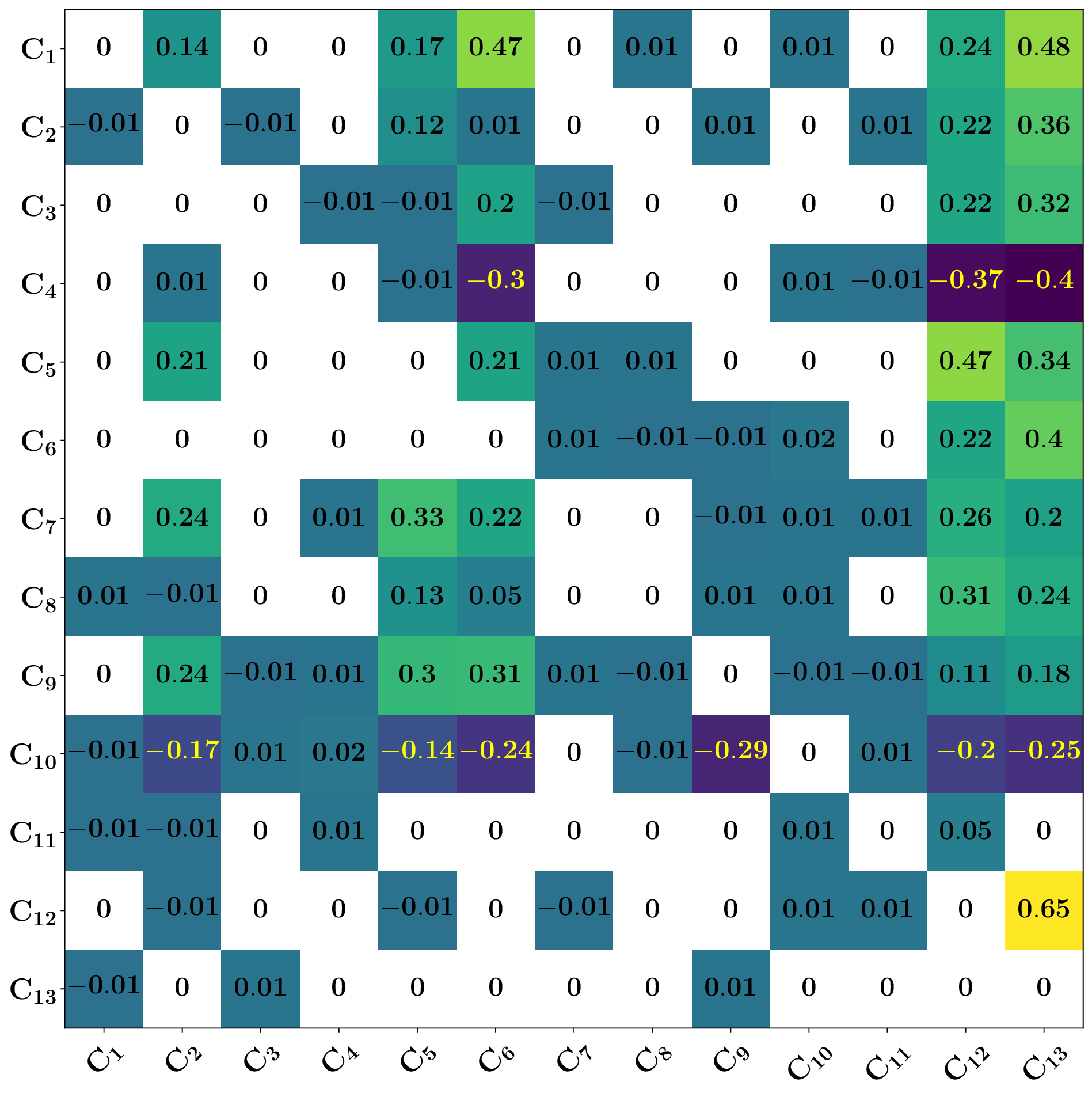}
 	\caption{Likelihood FCM mixture $\mathcal{F}_{\textsf{avg}}$}
    \label{fig:likelihood_Mixture}
    \end{subfigure}
    \hfill
   \begin{subfigure}[t]{0.72\textwidth}
    \centering
 	\includegraphics[width=\linewidth]{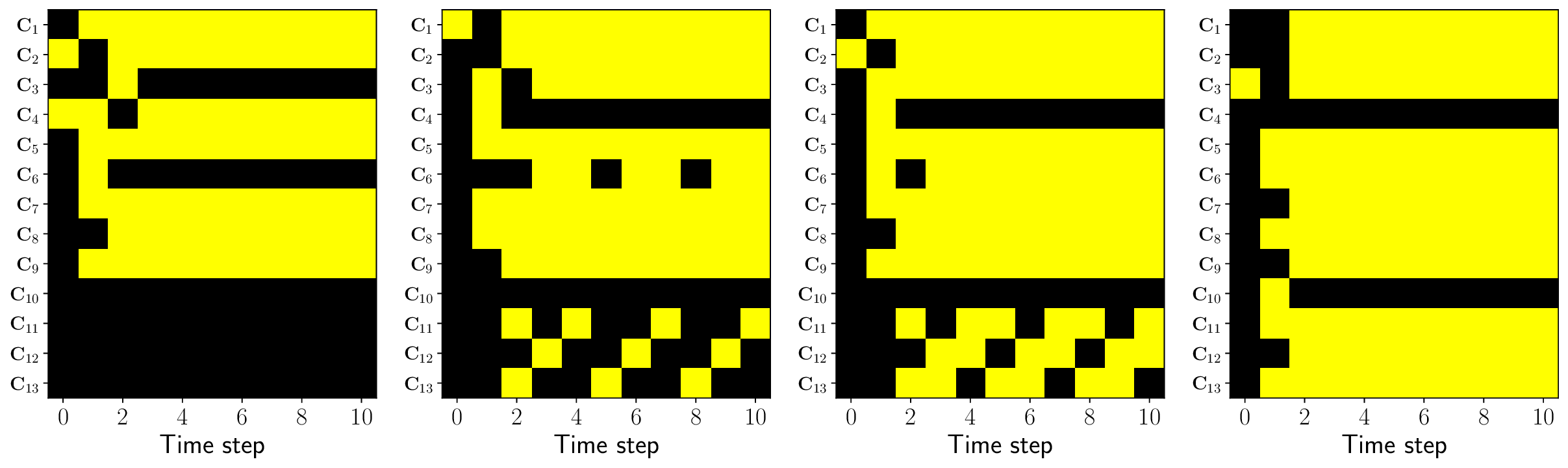}
 	\caption{ Attractors: Two fixed points and two limit cycles}
    \label{fig:likelihood_Mixture_Attractor}
    \end{subfigure}
   \caption{ \small Mixing the likelihood FCMs with uniform weights ($w_1=w_2 = w_3=w_4 = w_5 = w_6 = w_7 = \frac{1}{7}$): (a) The mixture $\mathcal{F}_{\textsf{avg}}$ averages the 7 likelihood FCM weight matrices and  produces a matrix whose entries blend the causal influences of the individual maps. (b) These are all the nontrivial equilibria for all the basins: two fixed points and two limit cycles. 
    }
 \end{figure*}

\begin{figure*}[htbp]
\centering
   \begin{subfigure}[t]{0.32\textwidth}
    \centering
 	\includegraphics[width=0.89\linewidth]{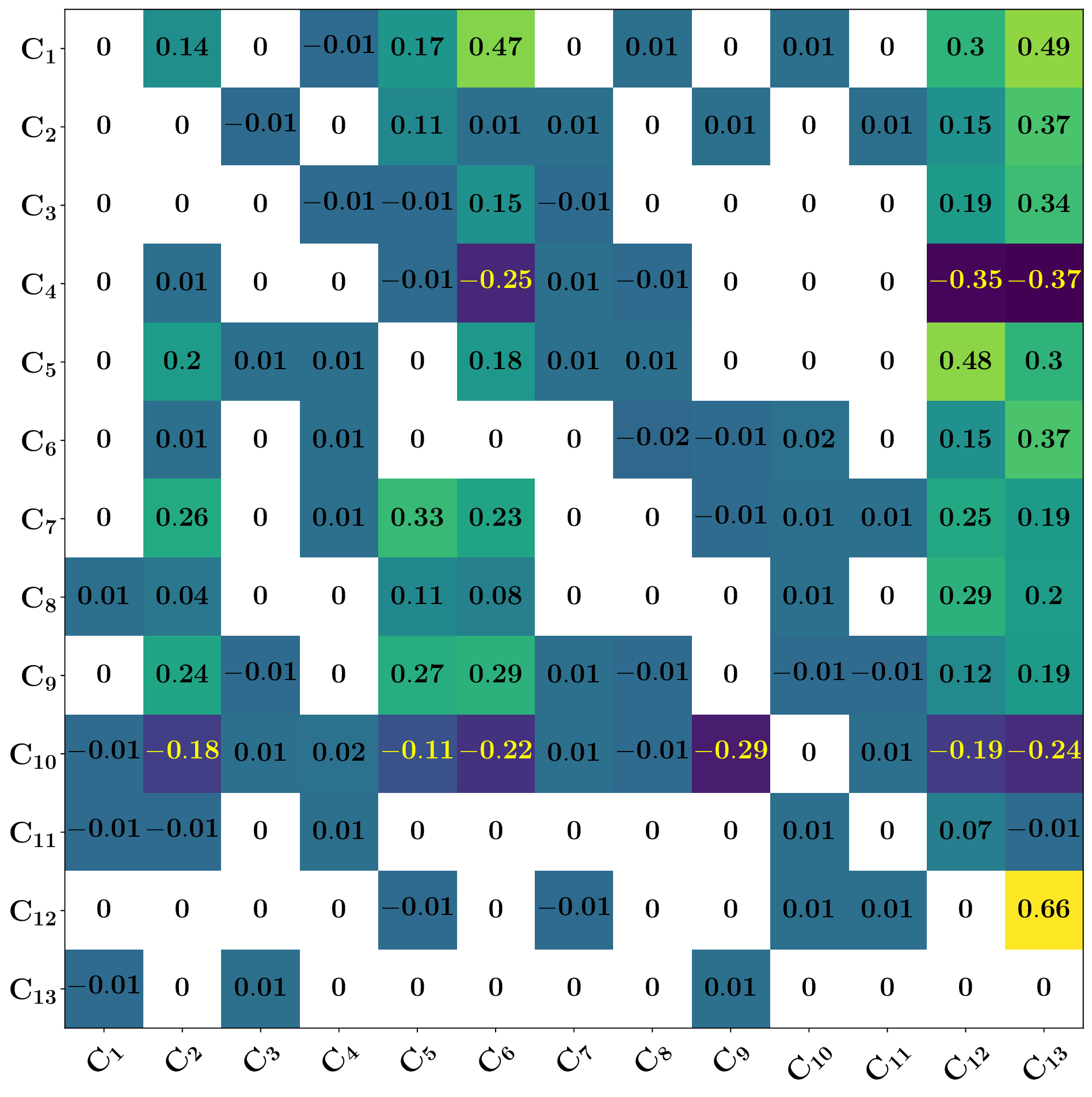}
 	\caption{Relevance FCM mixture with $x^{\prime}$}
    \label{fig:rel_fcm_weight_uniform_1}
    \end{subfigure}
    \begin{subfigure}[t]{0.32\textwidth}
    \centering
 	\includegraphics[width=0.9\linewidth]{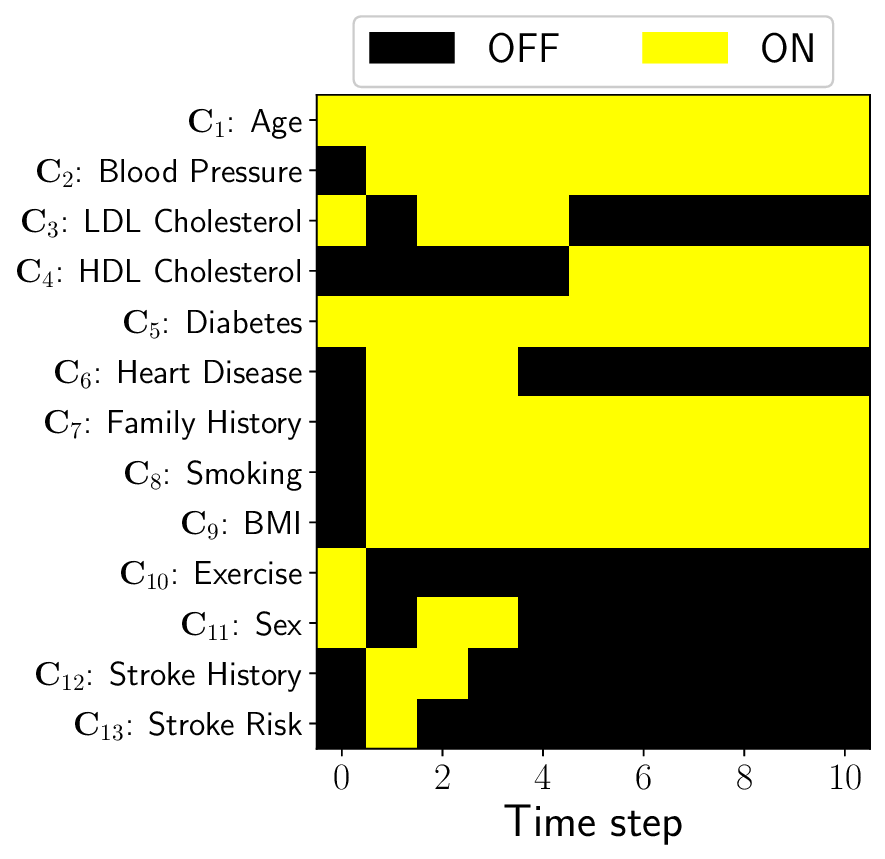}
 	\caption{Equilibrium: $\mathcal{F}_R$ and $x^{\prime}$}
    \label{fig:rel_fcm_att_uniform_x1}
    \end{subfigure}
    \begin{subfigure}[t]{0.32\textwidth}
    \centering
 	\includegraphics[width=0.9\linewidth]{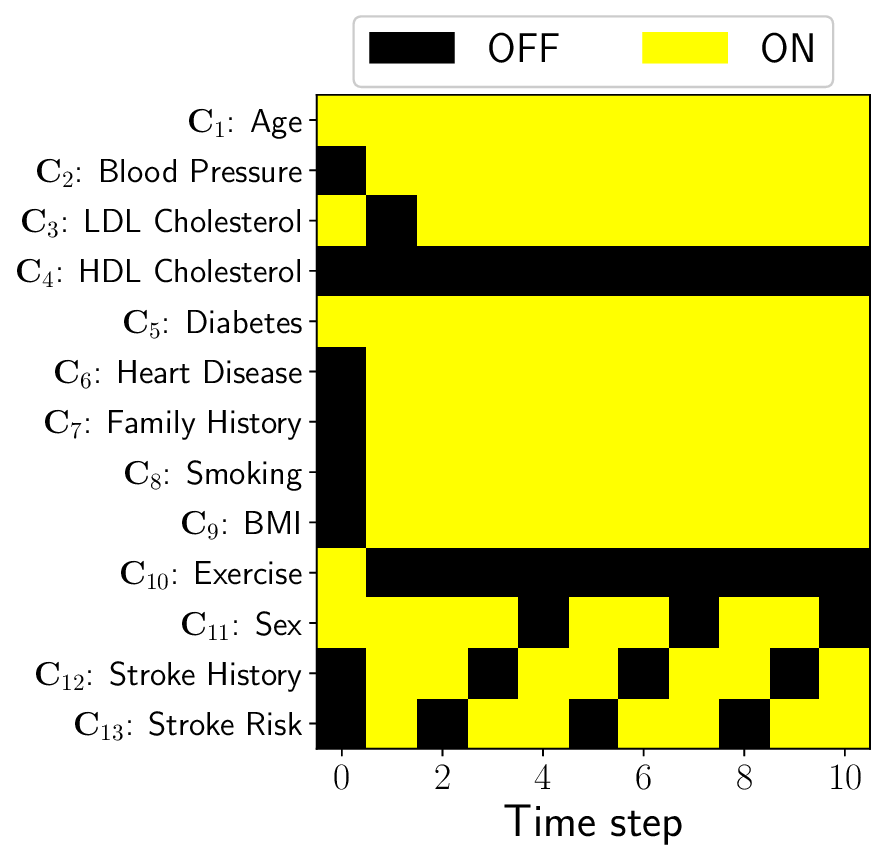}
 	\caption{Equilibrium: $\mathcal{F}_{\textsf{avg}}$ and $x^{\prime}$}
    \label{fig:mix_likelihood_fcm_att_uniform_x1}
    \end{subfigure}
    \hfill 
   \caption{ \small Relevance FCM mixture from the weighted likelihood mixture with the \emph{uniform weights} ($w_1 = w_2= ... = w_7 = \frac{1}{7}$): We used initial state $x^{\prime} = [1,0,1,0,1,0,0,0,0,1,1,0,0]$.
   % and $x^{\prime \prime} = [0,0,1,0,0,0,0,0,0,0,1,0,0]$:
  %(a) The edge matrix for the relevant mixture FCM $\mathcal{F}_R$ with the  initial state vector $x^{\prime}$.
  %(b)  The equilibrium for $x^{\prime}$ with the relevant mixture FCM $\mathcal{F}_R$. 
  %(c)  The equilibrium for  $x^{\prime}$ with the  likelihood FCM mixture $\mathcal{F}_{\textsf{avg}}$. 
  %The relevant FCM mixture converges to a different equilibrium from the likelihood FCM mixture.
 %(d) The edge matrix for the relevant mixture FCM $\mathcal{F}_R$ with the  initial state vector $x^{\prime \prime}$.
 %(e)  The equilibrium for $x^{\prime \prime}$ with the relevant mixture FCM $\mathcal{F}_R$. 
 %(f)  The equilibrium for $x^{\prime \prime}$ with the  likelihood FCM mixture $\mathcal{F}_{\textsf{avg}}$. 
  %The relevant FCM mixture converges to different equilibria from the likelihood FCM mixture.  
   The edge matrix of the relevant FCM mixture $\mathcal{F}_R$  is on the left. 
   The corresponding equilibrium is in the middle. 
   The plot on the right shows the corresponding equilibrium from the likelihood FCM mixture $\mathcal{F}_{\textsf{avg}}$ for the same input.
   %The relevant FCM mixture converges to different equilibria from the likelihood FCM mixture. 
   }
    \label{fig:relevant_fcm_mixture_with_uniform_wgts}
 \end{figure*}

%\begin{figure*}[htbp]
%\centering
 %  \begin{subfigure}[t]{0.40\textwidth}
 %  \centering
 %	\includegraphics[width=0.9\linewidth]{Figures/ischemic_weight_mix_likelihood_uniform_relevance_2.eps}
% 	\caption{Relevant FCM mixture}
 %   \label{fig:likelihood_mix2}
 %   \end{subfigure}
 %   \hfill 
%   \begin{subfigure}[t]{0.4\textwidth}
%    \centering
% 	\includegraphics[width=\linewidth]{Figures/ischemic_weight_mix_likelihood_uniform_relevance_att2.eps}
% 	\caption{ Attractor for  initial state}
 %   \label{fig:likelihood_mix2_attractor}
%    \end{subfigure}
%   \caption{ Relevant FCM mixture from the weighted likelihood mixture with \emph{uniform weights} and initial state $x = [1,0,1,1,0,0,0,0,0,0,0,0,0]$: 
 %   }
%    \label{fig:relevant_fcm_mixture_with_uniform_wgts_2}
% \end{figure*}
% \end{comment}

\begin{figure*}[!htbp]
\centering
   \begin{subfigure}[t]{0.32\textwidth}
    \centering
 	\includegraphics[width=0.89\linewidth]{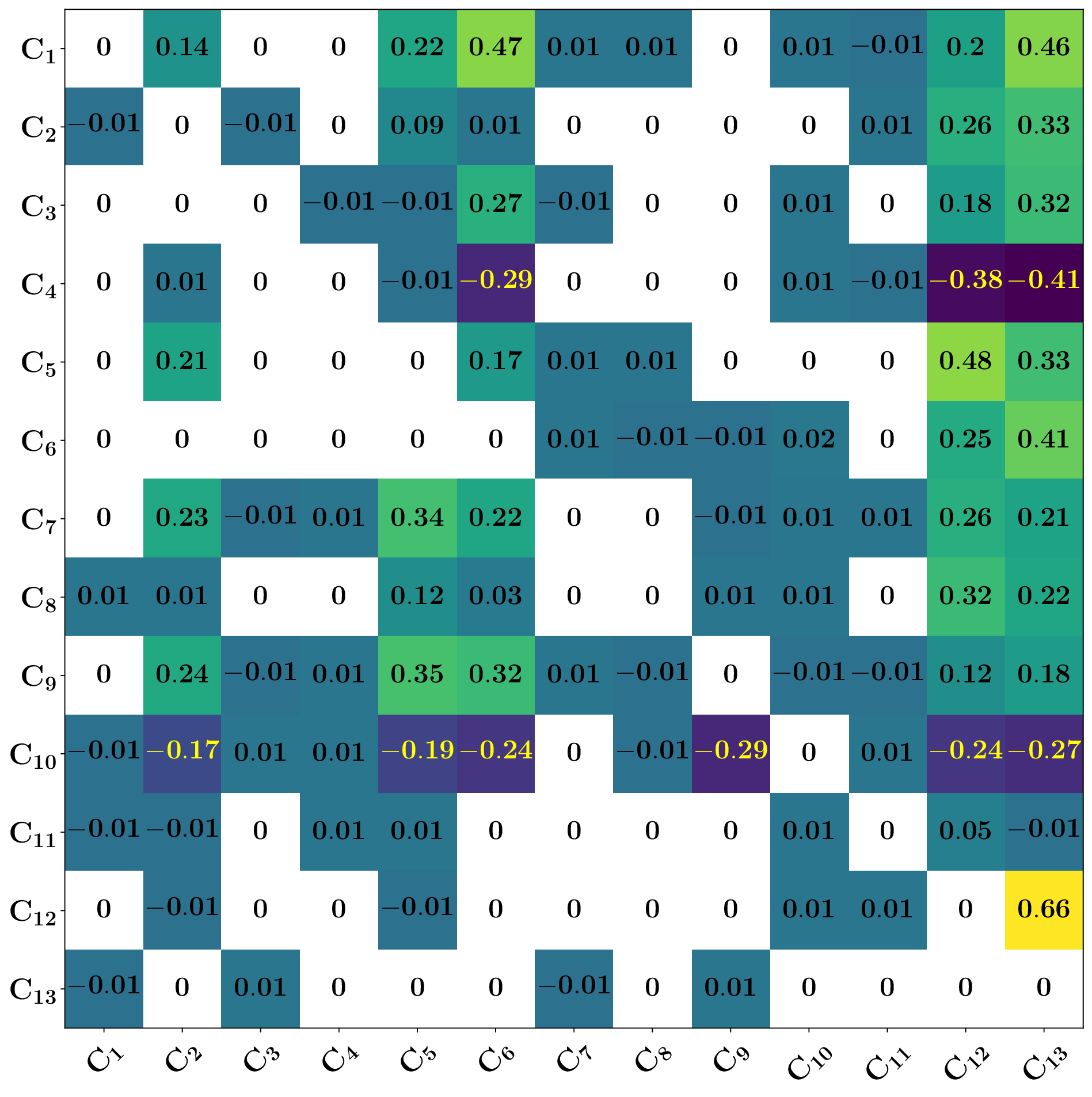}
 	\caption{Relevance FCM mixture with $x^{\prime \prime}$}
    \label{fig:rel_fcm_weight_nonuniform_1}
    \end{subfigure}
   \begin{subfigure}[t]{0.32\textwidth}
    \centering
 	\includegraphics[width=0.9\linewidth]{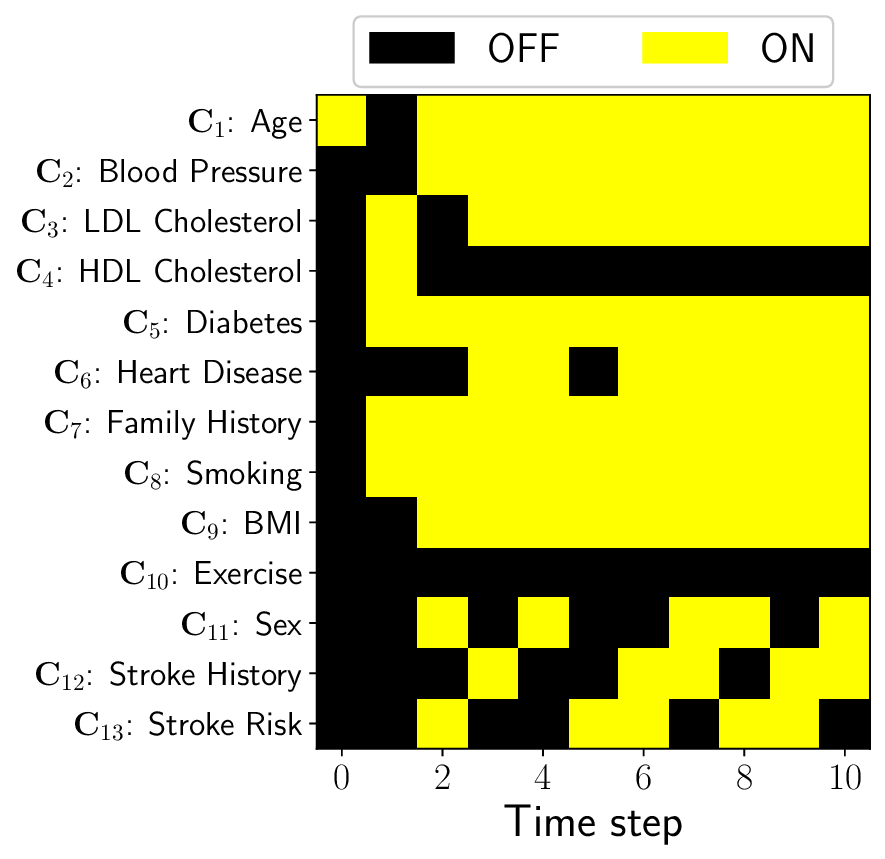}
 	\caption{Equilibrium: $\mathcal{F}_R$ and  $x^{\prime \prime}$}
    \label{fig:rel_fcm_att_nonuniform_x1}
    \end{subfigure}
        \begin{subfigure}[t]{0.32\textwidth}
    \centering
   	\includegraphics[width=0.9\linewidth]{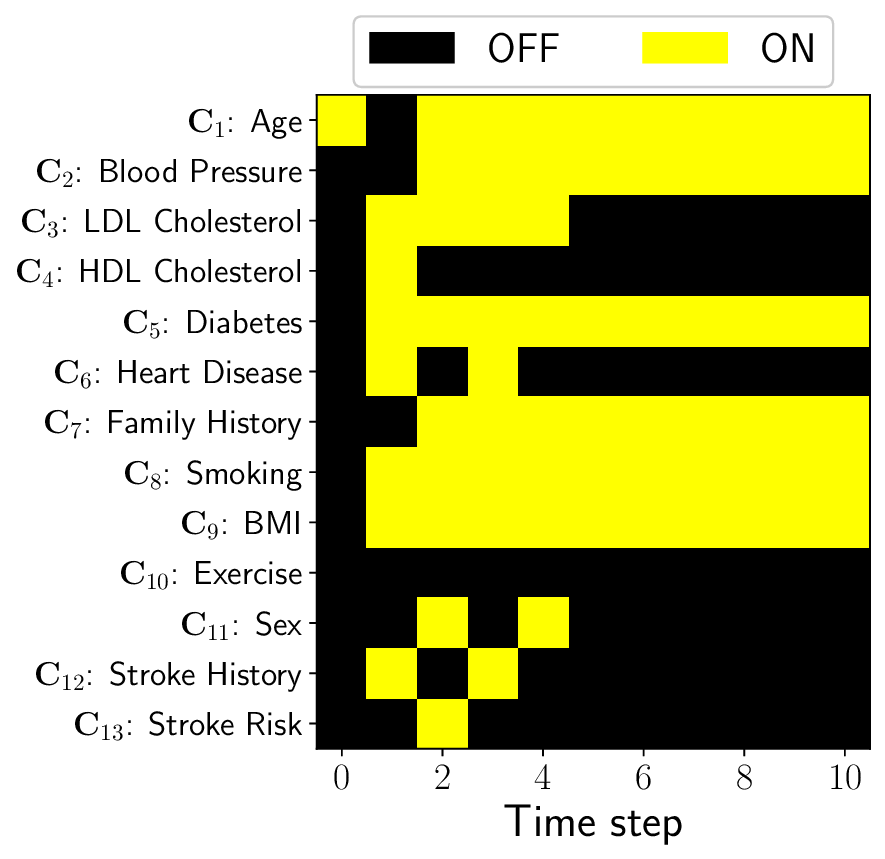}
   	\caption{Equilibrium: $\mathcal{F}_{\textsf{avg}}$ and $x^{\prime \prime}$}
    \label{fig:mix_likelihood_fcm_att_nonuniform_x1}
    \end{subfigure}
 %  \hfill 
 % \begin{subfigure}[t]{0.3\textwidth}
 %   \centering
 %  	\includegraphics[width=0.9\linewidth]{Figures/ischemic_weight_mix_likelihood_nonuniform_relevance_2.eps}
 %   	%\includegraphics[width=\linewidth]{Figures/ischemic_weight_mix_likelihood_nonuniform_relevance_att2.eps}
 %	%\includegraphics[width=\linewidth]{Figures/ischemic_weight_mix_likelihood_nonuniform_relevance_att1.eps}
 %	\caption{Relevance FCM mixture with $x^{\prime \prime}$}
 %   \label{fig:rel_fcm_weight_nonuniform_2}
  %  \end{subfigure} 
  %  \begin{subfigure}[t]{0.33\textwidth}
  %  \centering
  %  	%\includegraphics[width=0.9\linewidth]{Figures/ischemic_weight_mix_likelihood_nonuniform_relevance_2.eps}
  %  	\includegraphics[width=0.9\linewidth]{Figures/ischemic_weight_mix_likelihood_nonuniform_relevance_att2.eps}
  %	%\includegraphics[width=\linewidth]{Figures/ischemic_weight_mix_likelihood_nonuniform_relevance_att1.eps}
  %	\caption{Attractor: $\mathcal{F}_R$ and $x^{\prime \prime}$}
  %  \label{fig:rel_fcm_att_nonuniform_x2}
  % \end{subfigure}
  %  \begin{subfigure}[t]{0.33\textwidth}
  %  \centering
 %	%\includegraphics[width=0.9\linewidth]{Figures/ischemic_weight_mix_likelihood_uniform_relevance.eps}
 %	\includegraphics[width=0.9\linewidth]{Figures/ischemic_weight_mix_likelihood_nonuniform_initial_state_2.eps}
  %	\caption{Attractor: $\mathcal{F}_{\textsf{avg}}$ and $x^{\prime \prime}$}
    %\label{fig:mix_likelihood_fcm_att_nonuniform_x2}
   % \end{subfigure}
   \caption{\small Relevance FCM mixture from the weighted likelihood mixture with the \emph{non-uniform weights} ($w_1= 0.5, w_2 = 0.15,  w_3= 0.1, w_4 = 0.1,  w_5 =  0.05, w_6 = 0.05, w_7 = 0.05$): 
   We tested the initial state $x^{\prime \prime} = [1,0,0,0,0,0,0,0,0,0,0,0,0]$. %$x^{\prime \prime} = [0,0,0,0,1,1,0,0,0,0,0,0,0]$.  
   The edge matrix of the relevant FCM mixture $\mathcal{F}_R$  is on the left. 
   The corresponding equilibrium is in the middle. 
   The plot on the right shows the corresponding equilibrium from the likelihood FCM mixture $\mathcal{F}_{\textsf{avg}}$  for the same initial state.
   %The relevant FCM mixture converges to different equilibria from the likelihood FCM mixture.
    }
    \label{fig:relevant_fcm_mixture_with_nonuniform_wgts}
    %\vspace{-0.20in}
 \end{figure*}

\section{Simulations and Discussion}

\begin{table}[htbp]
\caption{\small Relevancy mixing weights for the posterior FCMs with the uniform weights  ($w_1=  w_2 = w_3 = w_4 = w_5 =  w_6 =  w_7 = \frac{1}{7}$).  We picked four initial states $x^{(i)}$'s from the four largest basins. }
\centering
\footnotesize
\begin{tabular}{|c|c|c|c|c|}
\hline
\multirow{2}{*}{Posterior $\mathcal{P}_j$} & \multirow{2}{*}{$r_j\big{(}x^{(1)}\big{)}$} & \multirow{2}{*}{$r_j\big{(}x^{(2)}\big{)}$} & \multirow{2}{*}{$r_j\big{(}x^{(3)}\big{)}$} & \multirow{2}{*}{$r_j\big{(}x^{(4)}\big{)}$} \\
&   &  &  & \\
\hline
$\mathcal{P}_1$   & $0.133$  &  \color{blue}{$\bf{0.175}$}  & $0.132$  & $0.101$  \\
$\mathcal{P}_2$   & $0.159$ &  $0.165$ & $0.155$  &  $0.159$ \\
$\mathcal{P}_3$   & $0.084$ &  $0.130$ &  $0.103$ &  $0.084$ \\
$\mathcal{P}_4$   & $0.133$ & $0.130$ &  $0.103$ &  $0.133$ \\
$\mathcal{P}_5$   & $0.168$ &  $0.154$  &  $0.161$ &  $0.139$ \\
$\mathcal{P}_6$   & \color{blue}{$\bf{0.178}$} & $0.130$ &  \color{blue}{$\bf{0.178}$} &  \color{blue}{$\bf{0.197}$} \\
$\mathcal{P}_7$   & $0.145$ & $0.116$  & $0.168$ &  $0.187$ \\
\hline
\end{tabular}
\label{tab:table_1}
%\vspace{-0.15in}
\end{table}

FCM perturbations generated 7 likelihood FCMs from the 13-node ischemic stroke FCM \cite{khodadadi2019fuzzy}. 
Figure~\ref{fig:ischemic_stroke_fcm_web} shows the ischemic stroke FCM. 
Figure~\ref{fig:ischemic_stroke_edge_wgt_matrix} shows its edge matrix and Figure~\ref{fig:ischemic_stroke_equilibria} shows its equilibria. 
It can converge to either a fixed point or a limit cycle. 

We used the noise perturbation method in Equation (\ref{eq:noise_pertubation}) to generate 7 synthetic likelihood FCMs with $c = 0.95$ and $d = 0.05$. 
The multinoulli trivalent mask from Equation (\ref{eq:multinouli_mask}) used $p=q=0.1$.
Figures~\ref{fig:likelihood_1}-\ref{fig:likelihood_7} show the edge matrices for the 7 likelihood FCMs and Figures~\ref{fig:likelihood_1_attractor}-\ref{fig:likelihood_7_attractor} show the respective equilibria for the 7 likelihood FCMs. 
These FCMs could each converge to 1 or 2 or even 3 equilibria. 
These equilibria could be fixed points or limit cycles. 

We then mixed the 7 likelihood FCMs according to the mixing weights $w_k$ to give the likelihood FCM mixture. 
Figure~\ref{fig:likelihood_Mixture} shows the edge matrix for an equal-weight mixture of the 7 likelihood FCMs:  $w_1=w_2=...=w_7=\frac{1}{7}$. 
Figure~\ref{fig:likelihood_Mixture_Attractor} shows the equilibria that equal-weight mixture converged to.
Every one of the possible  $2^{13}$ pulse inputs (binary vectors) converges to one of the four equilibria or to the trivial zero‑vector fixed point.
Figure~\ref{fig:likelihood_mix}  shows the edge matrix for a likelihood FCM mixture with a set of nonuniform  weights. 
Figure~\ref{fig:likelihood_mix_attractor}  shows the corresponding equilibrium.

The FCM mixture gives 7 posterior FCMs that then mix using the same mixing weights $w_k$ to give the posterior FCM mixture.
The posterior FCM equilibria and the equilibria from the posterior FCM mixture measured the relevance of the likelihood FCMs using the relevancy mixing weights $r_k(x)$. 
Table~\ref{tab:table_1} shows the relevancy weights for an equal-weight mixture of likelihood FCMs given several input states $x$.
We tested four cases of input states from the four largest basins of the corresponding posterior FCM mixture. 
We picked one vector per basin.
%The vectors are $x^{(1)} =$ [1,0,....,0], $x^{(2)} =$ [1,0,1,1,0,...,0], $x^{(3)} =$ [0,...,0,1,1,0,...,0], and  $x^{(4)}$ = [0,1,0,0,1,1,0,...,0].
Table~\ref{tab:table_2} shows the relevancy weights from a unequal-weight mixture of the likelihood FCMs for the same set of inputs $x$.
We also tested four cases of input states from the four largest basins of the corresponding posterior FCM mixture. 
We picked one vector per basin.
%The vectors are $x^{(1)} =$ [1,0,....,0], $x^{(2)} =$ [0,...,0,1,0,...,0], $x^{(3)} =$ [0,0,0,1,0,....,0], and  $x^{(4)}$ = [0,...,0,1,1,0,...,0].

The likelihood FCMs then mixed using mixing weights $r_k(x)$ to give the relevant FCM mixture $\mathcal{F}_R(x)$ that converged to its own equilibria.
%Figures~\ref{fig:rel_fcm_weight_uniform_1} and \ref{fig:rel_fcm_weight_uniform_2} show the edge matrices of the relevant FCM mixture for equal-weight likelihood FCM mixture for different input patterns. 
%Figures~\ref{fig:rel_fcm_att_uniform_x1} and \ref{fig:rel_fcm_att_uniform_x2} show the equilibria of the relevant FCM mixture for the two inputs. 
Figure~\ref{fig:rel_fcm_weight_uniform_1} shows the edge matrix of the relevant FCM mixture for the equal-weight likelihood FCM mixture for  input pattern $x^{\prime}$. 
Figure~\ref{fig:rel_fcm_att_uniform_x1} shows the equilibrium of the relevant FCM mixture for the input. 
The figure also compares these equilibria to those of the equal-weight likelihood FCM mixture given the same input in Figure~\ref{fig:mix_likelihood_fcm_att_uniform_x1}. %and \ref{fig:mix_likelihood_fcm_att_uniform_x2} 

Figure~\ref{fig:relevant_fcm_mixture_with_nonuniform_wgts} shows the relevant FCM mixture's edge matrix and equilibria for a set of unequal mixing weights $w_k$. 
This figure also compares the relevant FCM mixture's equilibria to those of the likelihood FCM mixture for the same input.
The input ``${\bf C}_1$: Age'' pushes the likelihood FCM mixture into a fixed point equilibrium where the node ``${\bf C}_{13}$: Stroke Risk'' is inactive. 
Figure~\ref{fig:mix_likelihood_fcm_att_nonuniform_x1} shows this equilibrium. 
This means that the likelihood FCM mixture predicts no risk of stroke when asked ``What happens when you age?''
The relevant FCM mixture converged to a 3-step limit cycle for the same input. 
Figure~\ref{fig:rel_fcm_att_nonuniform_x1} shows this equilibrium. 
The node ``${\bf C}_{13}$: Stroke Risk'' never \emph{stays} inactive. 
So there is an ever-present risk of stroke.

\section{Conclusion}
The relevance of an expert's answer depends on its probativeness and its materiality.
We model such relevance by how much the expert's FCM limit-cycle answer differs or varies from the limit-cycle answers of a likelihood-based or posterior-based mixture FCM.
The technique can control for subject-matter materiality or expert credibility to suppress or enhance large variations if needed.
This can also help produce relevant sparse mixtures.
New FCMs can easily add to the mixtures and do so recursively.
The computational cost depends on vector-matrix and pseudo-inverse operations on often sparse matrices.
So the technique is parallelizable and should admit hardware implementation for very large-scale knowledge processing.

\bibliographystyle{unsrt}

\bibliography{bibdata}

\end{document}